\RequirePackage{fix-cm}

\documentclass[twocolumn]{svjour3}          
\smartqed  
\usepackage{graphicx}
\usepackage{natbib}

\usepackage{tabulary,multirow,overpic}

\usepackage[misc,geometry]{ifsym} 
\usepackage{authblk}

\usepackage[colorlinks,citecolor=blue,urlcolor=blue,bookmarks=false,hypertexnames=true]{hyperref}
\usepackage[toc,page]{appendix}
\usepackage[table,dvipsnames]{xcolor}
\usepackage{arydshln}
\usepackage{amsmath,amsfonts,bm}
\usepackage{url}
\usepackage{mathtools}
\usepackage{amssymb}             
\usepackage{subcaption}
\usepackage{colortbl}
\usepackage{booktabs}
\usepackage{pifont}
\usepackage{diagbox}

\newcommand{\ie}{\textit{i}.\textit{e}.}
\newcommand{\eg}{\textit{e}.\textit{g}.}
\newcommand{\me}{\mathrm{e}}%

\definecolor{my_color1}{HTML}{8da0cb}
\definecolor{my_color2}{HTML}{66c2a5}

\newcommand{\xmark}{\ding{55}}%

\def\1{\bm{1}}

\DeclareMathOperator*{\argmin}{arg\,min}

\newcommand{\apbbox}[1]{AP$^\text{bb}_\text{#1}$}
\newcommand{\apmask}[1]{AP$^\text{mk}_\text{#1}$}

\journalname{International Journal of Computer Vision}

\begin{document}
\sloppy

\title{Slimmable Networks for Contrastive Self-supervised Learning
}

\author{Shuai Zhao
\and Linchao Zhu\thanks{Communicated by Linchao Zhu}
\and Xiaohan Wang
\and Yi Yang}

\institute{
$^{\textrm{\Letter}}$Linchao Zhu$^{1}$ \\
\email{zhulinchao@zju.edu.cn} \\
\\
Shuai Zhao$^{1,2*}$ \\
\email{zhaoshuaimcc@gmail.com}  \\
\\
Xiaohan Wang$^{1}$ \\
\email{wxh1996111@gmail.com}  \\
\\
Yi Yang$^{1}$ \\
\email{yangyics@zju.edu.cn} \\
\at
1 ReLER Lab, CCAI, Zhejiang University \\
2 ReLER Lab, AAII, University of Technology Sydney \\
* This work was done when Shuai is a research assistant at Zhejiang University.
}

\date{Received: date / Accepted: date}

\maketitle

\begin{abstract}
Self-supervised learning makes significant progress in pre-training large models, but struggles with small models.
Mainstream solutions to this problem rely mainly on knowledge distillation,
which involves a two-stage procedure: first training a large teacher model and then distilling it to improve the generalization ability of smaller ones.
In this work, we introduce another one-stage solution to obtain pre-trained small models without the need for extra teachers, namely, slimmable networks for contrastive self-supervised learning (\emph{SlimCLR}).
A slimmable network consists of a full network and several weight-sharing sub-networks,
which can be pre-trained once to obtain various networks,
including small ones with low computation costs.
However,
interference between weight-sharing networks leads to severe performance degradation in self-supervised cases,
as evidenced by \emph{gradient magnitude imbalance}
and \emph{gradient direction divergence}.
The former indicates that a small proportion of parameters produce dominant gradients during backpropagation,
while the main parameters may not be fully optimized.
The latter shows that the gradient direction is disordered, and the optimization process is unstable.
To address these issues, we introduce three techniques to make the main parameters produce dominant gradients and sub-networks have consistent outputs.
These techniques include slow start training of sub-networks, online distillation, and loss re-weighting according to model sizes.
Furthermore, theoretical results are presented to demonstrate that a single slimmable linear layer is sub-optimal during linear evaluation.
Thus a switchable linear probe layer is applied
during linear evaluation.
We instantiate SlimCLR with typical contrastive learning frameworks and achieve better performance than previous arts with fewer parameters and FLOPs.
The code is available at \url{https://github.com/mzhaoshuai/SlimCLR}.
\end{abstract}

\section{Introduction}
Over the last decade, deep learning achieves
significant success in various fields of
artificial intelligence,
primarily due to a significant amount of manually labeled data.
However, manually labeled data is expensive and far less available than unlabeled data in practice.
To overcome the constraint of costly annotations,
self-supervised learning
\citep{DosovitskiyFSRB16, wu2018unsupervised, cpcv1,2020_moco,2020_simclr}
aims to learn transferable representations
for downstream tasks by training networks on unlabeled data.
There has been significant progress in large models,
which are larger than ResNet-50~\citep{2016_ResNet} 
that has roughly 25M parameters.
For example, ReLICv2~\citep{relicv2} achieves an
accuracy of 77.1\% on ImageNet~\citep{ImageNet} under a linear evaluation protocol with ResNet-50,
outperforming the supervised baseline 76.5\%.

\begin{figure*}[t]
	\centering
	\includegraphics[width=0.48\linewidth]{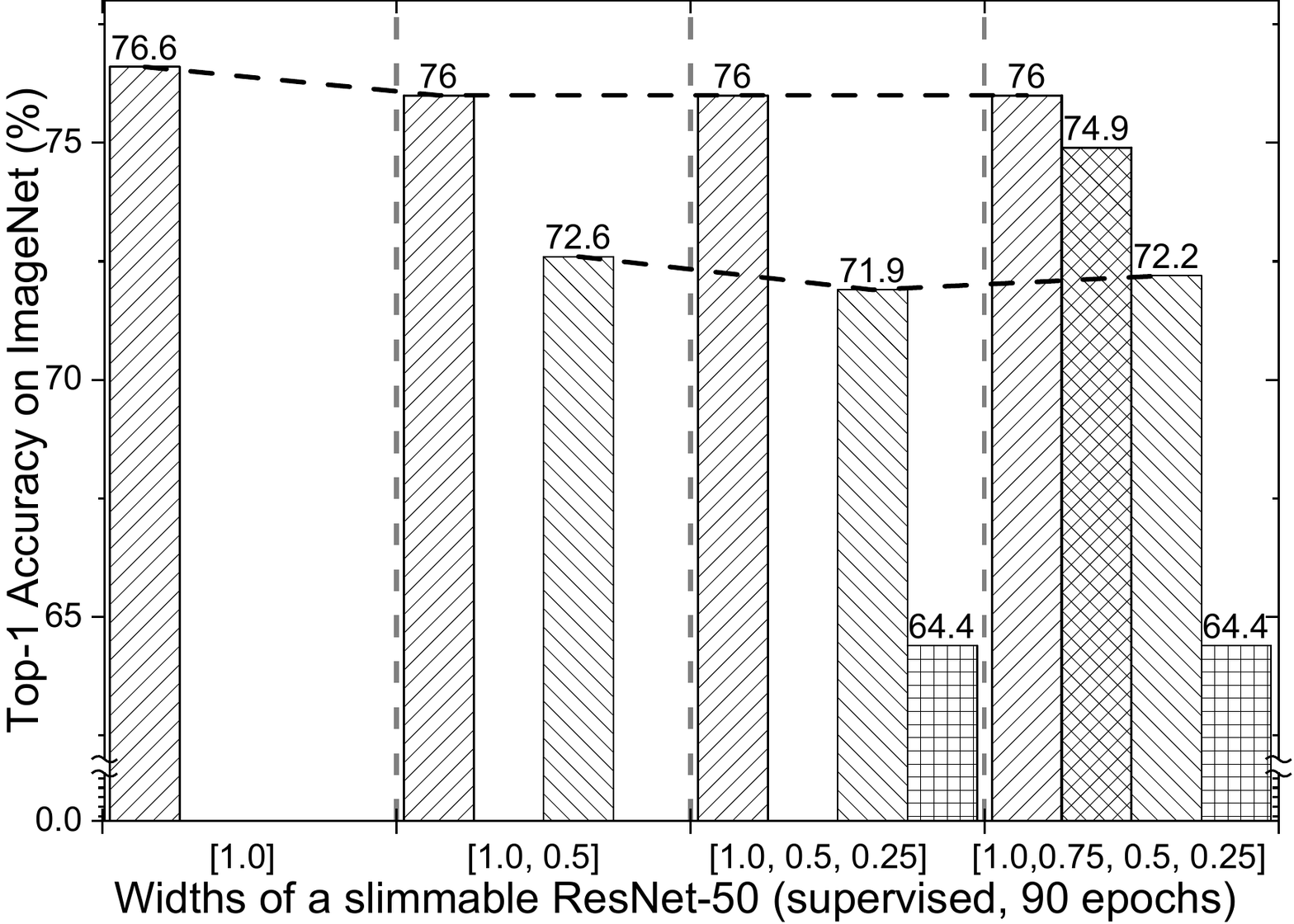}
	\includegraphics[width=0.48\linewidth]{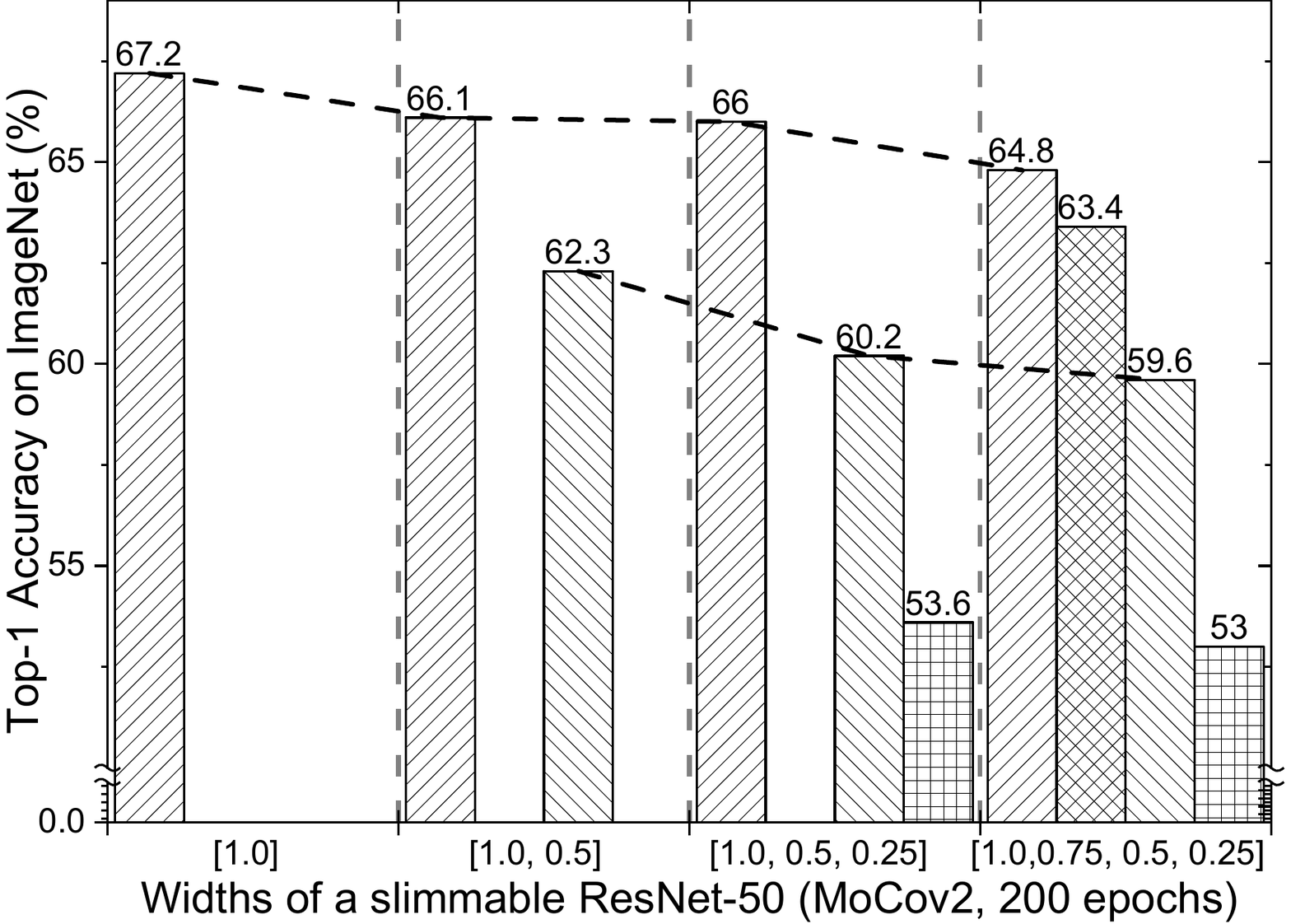}
	\caption{A slimmable ResNet-50 in supervised (left) and self-supervised (right) manners. $\text{ResNet-50}_{[1.0, 0.75, 0.5, 0.25]}$ means this slimmbale 
		network can switch at width $[1.0, 0.75, 0.5, 0.25]$. The width ${0.25}$ represents that the number of channels is scaled by $0.25$ of the full network.}
	\label{fig1_scales}
\end{figure*}

In contrast to the success of the large model pre-training,
self-supervised learning with small models lags behind.
For instance, supervised ResNet-18 with 12M parameters 
achieves an accuracy of 72.1\%  on ImageNet,
but its self-supervised result with MoCov2~\citep{mocov2} is only 52.5\%~\citep{seed}.
The gap is nearly 20\%.
To bridge the significant performance gap between supervised and self-supervised small models,
popular methods~\citep{compress,seed, disco, 2022_bingo} mainly focus
on knowledge distillation,
where they attempt to transfer the knowledge of a self-supervised large model into small ones.
Nevertheless, such methodology
has a two-stage procedure:
first train an additional large model,
and then train a small model to mimic the large one.
\cite{shi2022efficacy} investigate the impact of various components in contrastive self-supervised learning with small models to enhance performance without distillation signals.
They attain significant progress but are not superior to distillation methods.
Moreover, these methods only yield a single small model during pre-training, restricting the versatility of the pre-trained model in different resource-constrained scenarios.

An interesting question arises: \textit{can we obtain multiple small models through one-time pre-training to meet various computation scenarios without the need for additional teachers? }
Inspired by the success of slimmable networks
in supervised learning~\citep{2019_slim,2019_unslim,yu2019autoslim},
we develop a novel one-stage method for obtaining pre-trained small models without extra large models: slimmable
networks for contrastive self-supervised learning,
referred to as \emph{SlimCLR}.
A slimmable network comprises a full network and several weight-sharing sub-networks with different widths,
where the width represents the number of channels in a network.
The slimmable network can operate at various widths, allowing flexible deployment on different computing devices.
Therefore, we can obtain multiple networks, including small ones for low computing scenarios, through one-time pre-training.
Weight-sharing networks can also inherit knowledge from large ones via shared parameters, resulting in better generalization performance.

Weight-sharing networks in a slimmable network cause interference with each other when training simultaneously, particularly in self-supervised cases.
In Figure~\ref{fig1_scales},
weight-sharing networks only have a slight impact on each other under supervision, with the full model achieving 76.6\% 
\textit{vs}. 76.0\% accuracy in 
$\text{ResNet-50}_{[1.0]}$ and
$\text{ResNet-50}_{[1.0, 0.75, 0.5, 0.25]}$.
However, without supervision,
the corresponding numbers become 67.2\% \textit{vs}. 64.8\%.
The observed evidence of interference among weight-sharing networks includes
\emph{gradient magnitude imbalance}
and 
\emph{gradient direction divergence}.
Gradient magnitude imbalance occurs when a small proportion of parameters produce dominant gradients during backpropagation.
This happens because the shared parameters receive gradients from multiple losses of different networks,
which can result in the majority of parameters not being fully explored.
Gradient direction divergence refers to the situation where the gradient directions of the full network are disordered, resulting in an unstable optimization process.
In self-supervised cases,
weighting-sharing networks may produce diverse
outputs without strong constraints such as labels
in supervised cases,
leading to conflicts in gradient directions.

To alleviate gradient magnitude imbalance,
it is essential to ensure that the main parameters generate dominant gradients during backpropagation.
To avoid conflicts in gradient directions of
weight-sharing networks,
they should have consistent outputs.
To achieve these goals, we propose three
simple yet effective techniques during
pre-training to alleviate interference among networks.
1)
We adopt a \textit{slow start} strategy for sub-networks.
The networks and pseudo supervision of contrastive learning are both unstable and fast updating
at the start of training.
To avoid interference making the situation worse, we only train the full model initially.
After the full model becomes relatively stable,
sub-networks can inherit learned knowledge via shared parameters and begin with better initialization.
2)
We apply \textit{online distillation}
to make all sub-networks consistent
with the full model, thereby eliminating the gradient divergence of networks.
The predictions of the full model
serve as consistent guidance for all sub-networks.
3)
We \textit{re-weight the losses of networks}
according to their widths to ensure that the
full model dominates the optimization process.
Additional, theoretical results are provided
to demonstrate that a single slimmable linear
layer is sub-optimal during linear evaluation.
Therefore, we
adopt a \textit{switchable linear probe layer} to
avoid the interference caused by parameter-sharing.

We instantiate two algorithms for SlimCLR with typical
contrastive learning frameworks, \ie, MoCov2~\citep{mocov2}
and MoCov3~\citep{2021_mocov3}.
Extensive experiments on ImageNet~\citep{ImageNet} show that SlimCLR achieves significant performance improvements compared to previous arts with fewer parameters and FLOPs.
\section{Related Works}
\paragraph{\textbf{Self-supervised learning}}
Self-supervised learning aims to learn transferable representations
for downstream tasks from the input data itself.
According to~\cite{2021_survey},
self-supervised methods can be summarized into three main categories
according to their objectives:
\textit{generative}, \textit{contrastive},
and \textit{generative-contrastive (adversarial)}.
Methods belonging to the same categories can be further 
classified by the difference between pretext tasks.
Given input $x$, generative methods encode $x$ into an explicit vector
$z$ and decode $z$ to reconstruct $x$ from $z$,
\eg, auto-regressive~\citep{2016_pixelcnn,2016_pixelrnn},
auto-encoding models~\citep{1987_ae,2014_VAE,2019_bert,2022_mae}.
Contrastive learning methods encoder input $x$ into an explicit vector $z$
to measure similarity.
The two mainstream methods below this category are context-instance contrast
(infoMax~\cite{2019_infomax}, CPC~\cite{cpcv1},
AMDIM~\cite{2019_addim})
and instance-instance contrast
(DeepCluster~\cite{2018_deepcluster}, 
MoCo
~\cite{2020_moco,2021_mocov3},
SimCLR
~\cite{2020_simclr,2020_simclrv2}, 
SimSiam~\cite{2021_simsiam}).
Generative-contrastive methods generate a fake 
sample $x^\prime$ from $x$
and try to distinguish $x^\prime$ from real 
samples, \eg,
DCGANs~\cite{2016_dcgans},
inpainting~\cite{2016_inpaint}, and
colorization~\cite{2016_color}.

\paragraph{\textbf{Self-supervised small models}}
While self-supervised learning has made significant progress with large models like ResNet-50~\citep{2021_mocov3,relicv2}, small models struggle with common self-supervised pretext tasks.
There exists a notable performance gap between self-supervised small models and their supervised counterparts in downstream tasks.
To address this, typical solutions employ knowledge distillation~\citep{2015_kd}, which involves incorporating an additional self-supervised large model to guide the pre-training of the small model.
The key lies in aligning the predicted distributions of the student and the teacher, which arise from different data views of the same sample.
Alignment is generally achieved by minimizing their Kullback–Leibler~(KL) divergence~\citep{compress,seed}.
SimDis~\citep{gu2021simple} and DisCo~\citep{disco} substitute the KL divergence with $\ell_2$ distance. 
Additionally, to enhance generalization, BINGO~\citep{2022_bingo} transfers relationships among similar samples produced by the teacher to the predictions of the student.
Beyond distillation methods, \cite{shi2022efficacy} improve the performance of self-supervised small models by carefully tuning various components of contrastive self-supervised learning.
While they achieve considerable progress, their results does not surpass those of distillation methods.

\begin{figure*}[!t]
	\centering
	\begin{subfigure}{0.94\linewidth}
		\includegraphics[width=\linewidth]{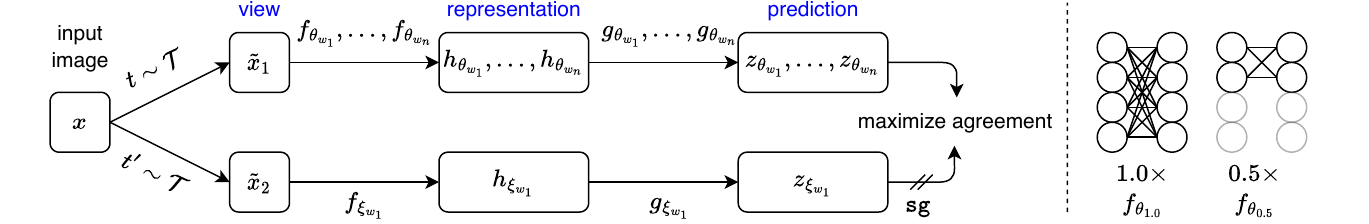}
		\caption{left: SlimCLR-MoCov2. right: a slimmable network with two widths $[1.0, 0.5]$.}
		\label{fig2:sub0}
	\end{subfigure}
	\begin{subfigure}{0.94\linewidth}
		\includegraphics[width=\linewidth]{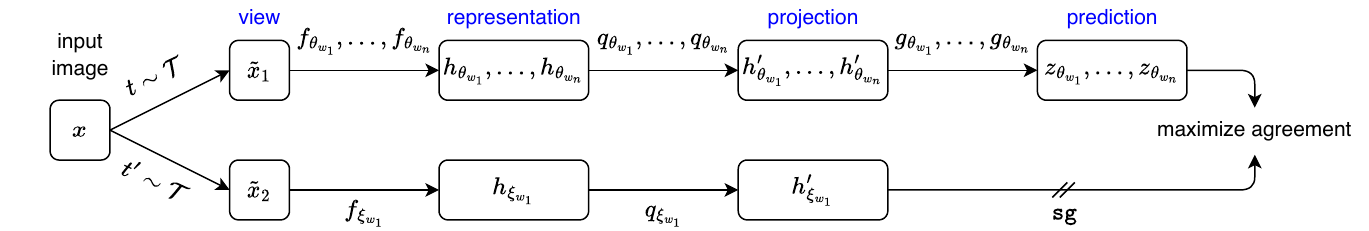}
		\caption{SlimCLR-MoCov3}
		\label{fig2:sub1}
	\end{subfigure}	
 	\caption{
 	The overall framework of SlimCLR.
		A slimmable network produces different outputs from weight-sharing networks with various widths
		$w_1,\ldots,w_n$, where $w_1$ is the width of the full model.
		$\theta$ are the network parameters and $\xi$
		are an exponential moving average of $\theta$. $\mathtt{sg}$ means stop-gradient.
	}
	\label{fig2_framework}
\end{figure*}

\paragraph{\textbf{Slimmable networks}}
Slimmable networks are first proposed to achieve instant and adaptive 
accuracy-efficiency trade-offs on different devices~\citep{2019_slim}.
A slimmable network can execute at different widths during runtime.
Following the pioneering work, universally slimmable networks
\citep{2019_unslim} develop
systematic training approaches to allow slimmable networks
to run at arbitrary widths.
AutoSlim~\citep{yu2019autoslim} further achieves one-shot architecture
search for channel numbers under a certain computation budget.
MutualNet~\citep{2020_mutualnet} trains slimmable networks
using different input resolutions to learn multi-scale representations.
Dynamic slimmable networks~\citep{Li_2022_pami,Li_2021_CVPR}
change the number of channels of each layer on the fly according to the input. 
In contrast to weight-sharing sub-networks in slimmable networks, some methods
aim to train multiple sub-networks with independent parameters~\citep{2020_dc}.
A relevant concept of slimmable networks in network pruning is
\textit{network slimming} \citep{2017_networkslim,2022_vitslim,pmlr-v139-wang21e}, which aims to 
achieve channel-level sparsity for better computation efficiency.

\paragraph{\textbf{Knowledge distillation}}
Knowledge distillation~\citep{2015_kd} aims to transfer the knowledge of a large network to a small one, and can be roughly categorized into two types:
logits distillation~\citep{2015_kd,2018_dml,guo2022reducing,zhao2022decoupled,2020_improved_kd} and
intermediate feature
distillation~\citep{2019_relationalkd,2020_CRD,2019_CCKD}.
The former only requires the student to mimic the output of the teacher, while the latter also aligns the intermediate features of the
teacher and student.
Intermediate feature distillation methods generally achieve better performance than
logits distillation methods.
Recently, several methods~\citep{guo2022reducing,zhao2022decoupled} try to find out the limit of logits distillation and improve its performance.
In this paper, online distillation is a kind
of logits distillation. Besides,
in a slimmable network, distillation also
occurs between a large network and a small
network via the shared parameters.
\section{Method}
\subsection{Description of SlimCLR}
We develop two instantial algorithms for SlimCLR
with typical contrastive learning frameworks MoCov2 and MoCov3~\citep{mocov2,2021_mocov3}.
As shown in Figure~\ref{fig2:sub0} (right),
a slimmable network with $n$ widths 
${w_1},\ldots, {w_n}$
contains multiple weight-sharing networks $f_{\theta_{w_1}},\ldots, f_{\theta_{w_n}}$, which are parameterized by learnable weights 
${\theta_{w_1}},\ldots, {\theta_{w_n}}$,
respectively.
Each network $f_{\theta_{w_i}}$ in the slimmable network has its own set of weights $\Theta_{w_i}$ and $\theta_{w_i} \in \Theta_{w_i}$.
A network with a small width shares the weights
with large ones, namely,
$\Theta_{w_j} \subset \Theta_{w_i}$ if $w_j < w_i$.
Generally, we assume $w_j < w_i$ if $j > i$, \ie,
$w_1,\ldots,w_n$
arrange in descending order, and $\theta_{w_1}$ represent the parameters of the full model.

We first illustrate the learning process of 
SlimCLR-MoCov2 in Figure~\ref{fig2:sub0}.
Given a set of images $\mathcal{D}$,
an image $x$ sampled uniformly from $\mathcal{D}$,
and one distribution of image augmentation $\mathcal{T}$,
SlimCLR produces two data views $\tilde{x}_1 = t(x)$ and $\tilde{x}_2 = t^\prime(x)$
from $x$ by applying augmentations
$t \sim \mathcal{T}$ and $t^\prime \sim \mathcal{T}$, respectively.
For the first view, SlimCLR outputs multiple representations
$h_{\theta_{w_1}},\ldots, h_{\theta_{w_n}}$
and predictions
$z_{\theta_{w_1}},\ldots, z_{\theta_{w_n}}$
~\footnote{In contrast to MoCov2 and SimCLR, where the output of the model $z_{\theta}$ is referred to as the projection, in this work, we refer to the final output of the model as the prediction. This is to maintain consistency with the notation used in SlimCLR-MoCov3 and to simplify the formulas.},
where $h_{\theta_{w_i}} = f_{\theta_{w_i}}(\tilde{x}_1)$
and $z_{\theta_{w_i}} = g_{\theta_{w_i}}(h_{\theta_{w_i}})$.
$g$ is a stack of slimmable linear transformation layers, \ie,
a slimmable version of the MLP head in MoCov2 and SimCLR~\citep{2020_simclr}.
For the second view, SlimCLR only outputs a single representation from the full model
$h_{\xi_{w_1}} = f_{\xi_{w_1}}(\tilde{x}_2)$
and prediction
$z_{\xi_{w_1}} = g_{\xi_{w_1}}(h_{\xi_{w_1}})$.
We minimize the InfoNCE~\citep{cpcv1} loss with respect to $\theta_{w_i}$ to maximize the similarity of
positive pairs $z_{\theta_{w_i}}$ and $z_{\xi_{w_1}}$:
\begin{align} \label{infonce}
	\mathcal{L}_{\theta_{w_i}} =
 -\log \frac{ \me^{\overline{z}_{\xi_{w_1}}\cdot 
 \frac{\overline{z}_{\theta_{w_i}} }{\tau_1}}
 }{\me^{\overline{z}_{\xi_{w_1}}\cdot
 \frac{\overline{z}_{\theta_{w_i}}}{\tau_1}}
  + \sum_{z^-}\me^{z^-\cdot
 \frac{\overline{z}_{\theta_{w_i}}}{\tau_1}}},
\end{align}
where  $\overline{z}_{\theta_{w_i}}=z_{\theta_{w_i}} / \lVert z_{\theta_{w_i}}\rVert_2$,
$\overline{z}_{\xi_{w_1}}=z_{\xi_{w_1}} / \lVert z_{\xi_{w_1}}\rVert_2$,
$\tau_1$ is a temperature hyper-parameter,
and $\{z^-\}$ are features of negative samples.
In SlimCLR-MoCov2, $\{z^-\}$ are obtained from a queue which is updated every iteration during training by $\overline{z}_{\xi_{w_1}}$,
following the approach used in MoCov2.
Without any regularization during training,
the overall objective is the sum of losses of
all networks with various widths:
\begin{align}
	\mathcal{L}_\theta = \sum_{i=1}^{n}\mathcal{L}_{\theta_{w_i}}.
\end{align}
Here $\xi$ is updated by $\theta$ every iteration using a momentum coefficient
$m\in [0, 1)$, as follows:
$\xi \leftarrow m \xi + (1-m)\theta$.

Compared to SlimCLR-MoCov2, SlimCLR-MoCov3 has an
additional projection process.
Firstly, it projects the representation to another high dimensional space, then makes predictions.
The projector $q$ is a stack of slimmable 
linear transformation layers.
SlimCLR-MoCov3 also adopts the InfoNCE loss, but the
negative samples come from other samples in the mini-batch.

After contrastive learning, we only keep $f_{\theta_{w_1}},\ldots, f_{\theta_{w_n}}
$ and abandon other components.

\begin{figure*}[t]
	\centering
	\begin{subfigure}{.328\linewidth}
		\centering
		\includegraphics[width=\linewidth]{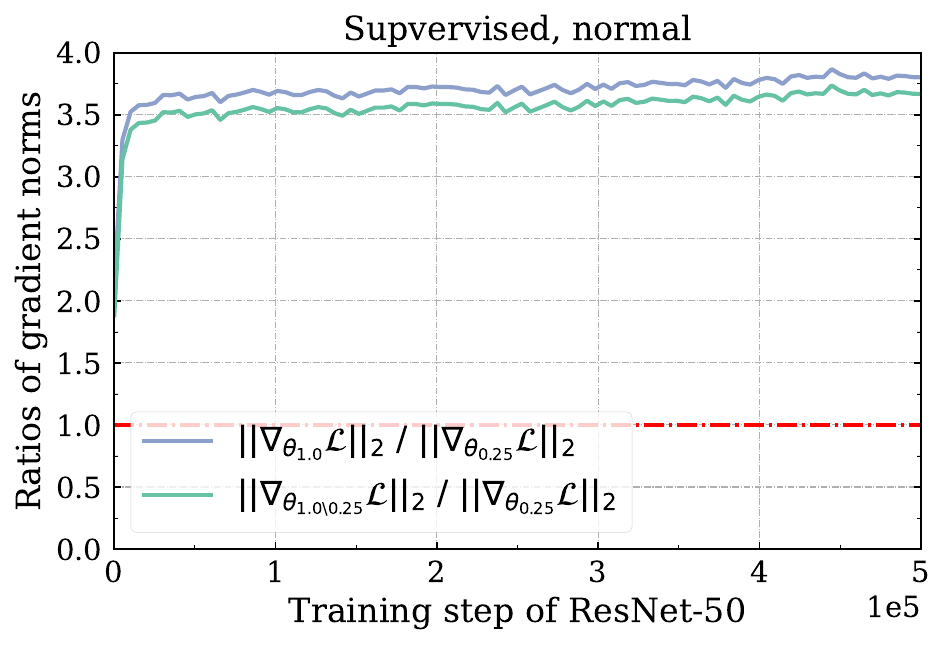}
		\caption{76.6\%}
		\label{fig3:sub0}
	\end{subfigure}%
	\begin{subfigure}{.328\linewidth}
		\centering
		\includegraphics[width=\linewidth]{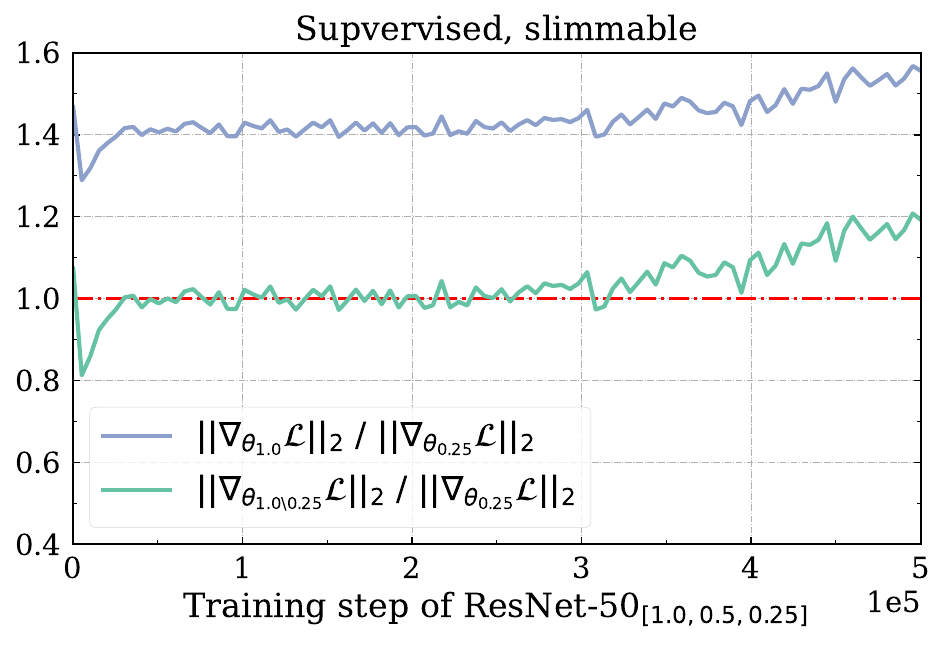}
		\caption{76.0\%, 72.9\%, 64.4\%}
		\label{fig3:sub1}
	\end{subfigure}%
	\begin{subfigure}{.328\linewidth}
		\centering
		\includegraphics[width=\linewidth]{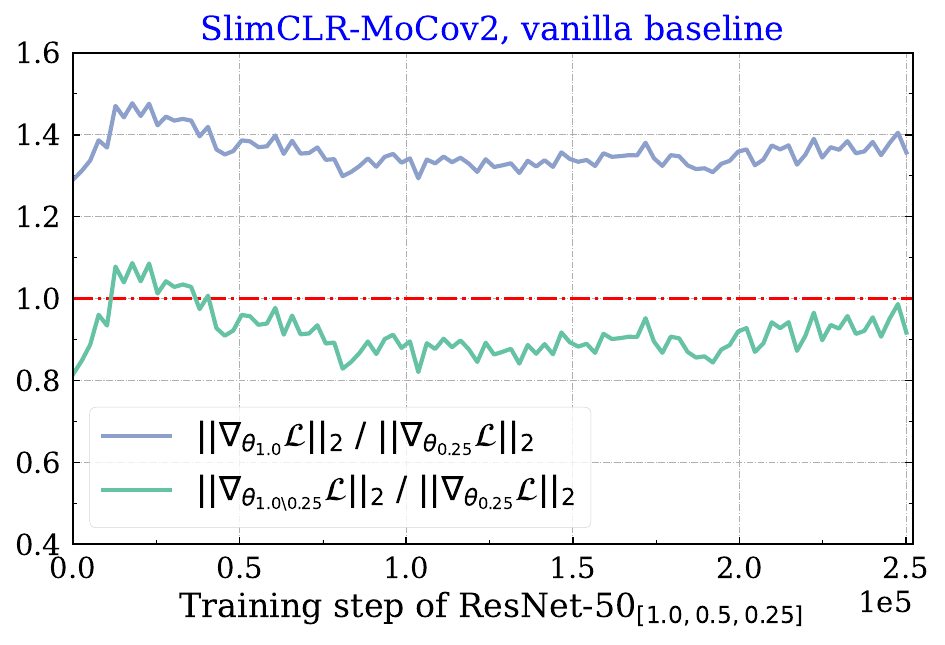}
		\caption{66.0\%, 61.4\%, 54.5\%}
		\label{fig3:sub2}
	\end{subfigure} \\
	\begin{subfigure}{.328\linewidth}
		\centering
		\includegraphics[width=\linewidth]{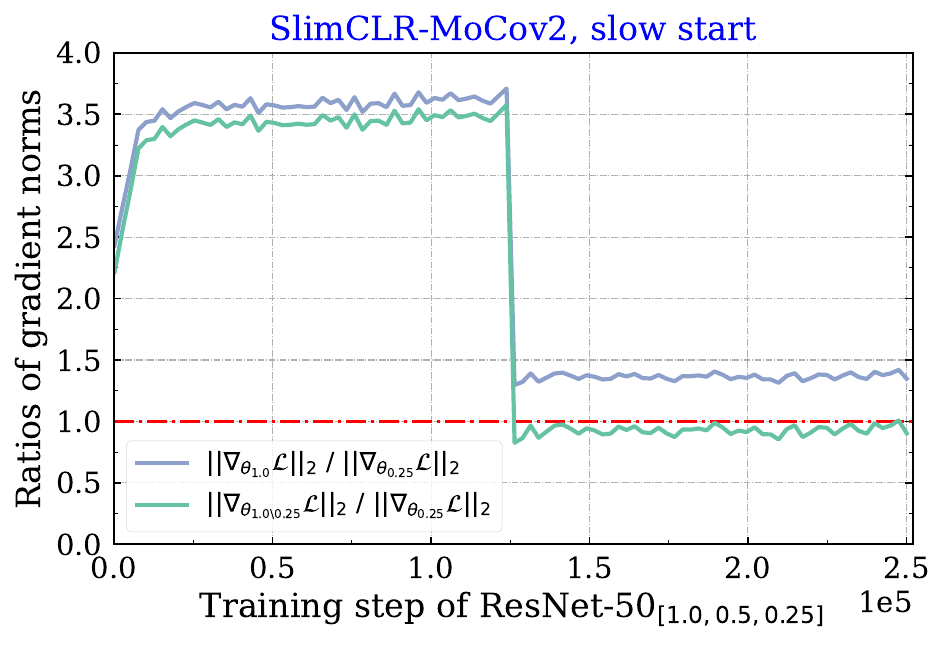}
		\caption{66.3\%, 62.2\%, 55.1\%}
		\label{fig3:sub3}
	\end{subfigure}
	\begin{subfigure}{.328\linewidth}
		\centering
		\includegraphics[width=\linewidth]{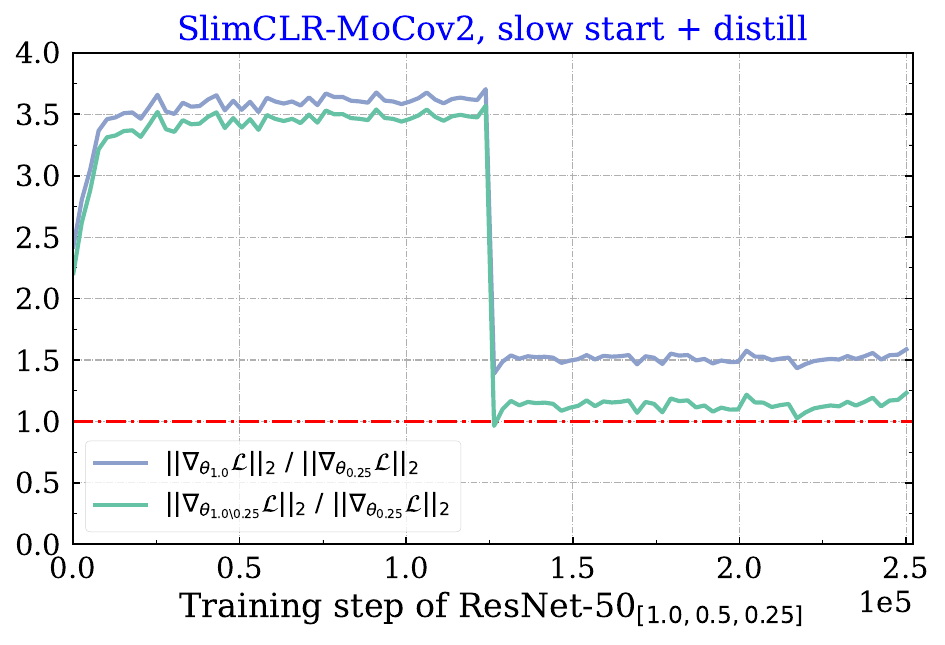}
		\caption{66.9\%, 62.3\%, 54.9\%}
		\label{fig3:sub4}
	\end{subfigure}
	\begin{subfigure}{.328\linewidth}
		\centering
		\includegraphics[width=\linewidth]{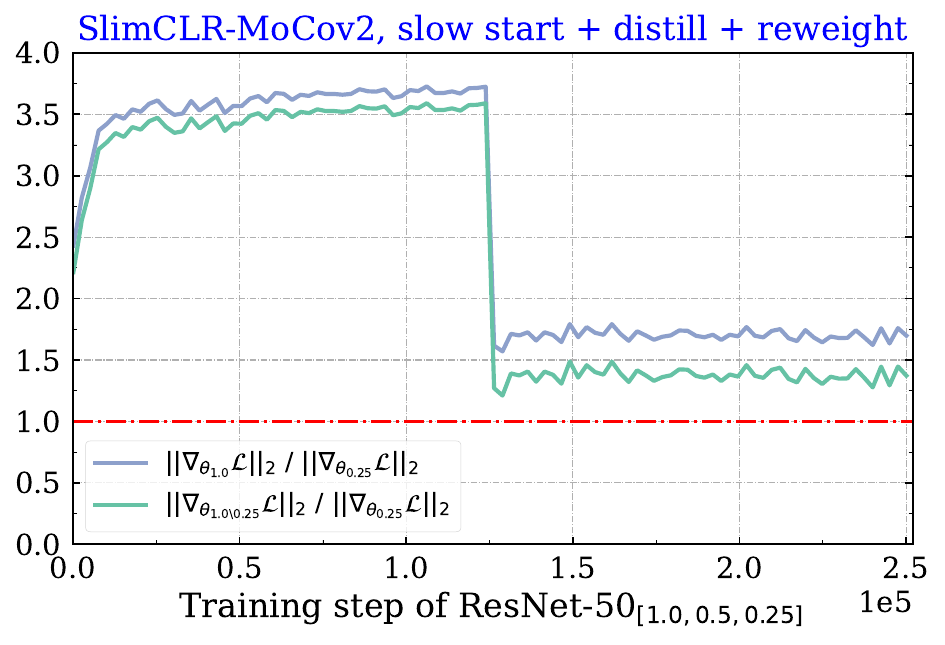}
		\caption{67.0\%, 62.4\%, 55.1\%}
		\label{fig3:sub5}
	\end{subfigure}
	\caption{
    Ratios of gradient norms:
    \textcolor{my_color1}{$\frac{\lVert \nabla_{\theta_{1.0}}\mathcal{L}_{\theta} \rVert_2}{\lVert\nabla_{\theta_{0.25}}\mathcal{L}_{\theta} \rVert_2}$}
    and
    \textcolor{my_color2}{$\frac{\lVert \nabla_{\theta_{1.0 \backslash 0.25}} \mathcal{L}_{\theta} \rVert_2}{ \lVert \nabla_{\theta_{0.25}} \mathcal{L}_{\theta} \rVert_2}$}.
    The gradient norm of the network is calculated by
    averaging the layer-wise $\ell_2$ gradient norms. $\nabla_{\theta_{1.0 \backslash 0.25}} \mathcal{L}_{\theta}$ is the
gradient of the final loss \emph{w.r.t.}
parameters $\theta_{1.0 \backslash 0.25} \in \Theta_{1.0} \backslash \Theta_{0.25}$, \ie, rest parameters of $\Theta_{1.0}$ besides $\Theta_{0.25}$.
}
	\label{fig3-grad}
\end{figure*}

\subsection{Interference of networks and solutions} \label{sec-grad}
In Figure~\ref{fig1_scales}, it can be observed that a vanilla implementation of SlimCLR suffers from significant performance degradation due to interference of weight-sharing networks. In this section, we will discuss the consequences of this interference and present solutions to address it.

\subsubsection{Gradient magnitude imbalance}
Gradient magnitude imbalance refers to the phenomenon where a small fraction of parameters receives dominant gradients during backpropagation.
For example,
in a slimmable network with widths $[1.0, 0.5, 0.25]$,
$\ell2$ norm of $\theta_{0.25}$ ---
$\lVert \nabla_{\theta_{0.25}} \mathcal{L}_{\theta} \rVert_2$
maybe larger than that of the remaining parameters 
$\lVert \nabla_{\theta_{1.0 \backslash 0.25}} \mathcal{L}_{\theta} \rVert_2$,
even though $\theta_{0.25}$ only counts for approximately $6\%$ of the total parameters.
This is because the gradients of different losses accumulate as follows:
\begin{align}
\nabla_{\theta_{w_n}} \mathcal{L}_{\theta}
	=\frac{\partial \mathcal{L}_{\theta}}{\partial \theta_{w_n}}
	=\sum_{i=1}^{n}\frac{\partial \mathcal{L}_{\theta_{w_i}}}{\partial {\theta_{w_n}}}.
\end{align}
Assuming that the gradients of a single loss have a similar magnitude distribution,
the accumulation of gradients from different losses increases the gradient norm of the shared parameters.

We evaluate the gradient magnitude imbalance by calculating the ratios of gradient norms during training.
Figure~\ref{fig3-grad} shows
the ratios of
$\lVert \nabla_{\theta_{1.0}} \mathcal{L}_{\theta} \rVert_2$
and $\lVert \nabla_{\theta_{1.0 \backslash 0.25}} \mathcal{L}_{\theta} \rVert_2$
versus $\lVert \nabla_{\theta_{0.25}} \mathcal{L}_{\theta} \rVert_2$,
separately.
The ratio of their numbers of parameters is
$\frac{\lvert{\Theta_{1.0} \backslash \Theta_{0.25}}\rvert}{\lvert{\Theta_{0.25}}\rvert}\approx 15$,
where $\theta_{1.0 \backslash 0.25} \in \Theta_{1.0} \backslash \Theta_{0.25}$.
In Figure~\ref{fig3:sub0},
both ratios of gradient norms are around 3.5, indicating that the majority of parameters obtain a large gradient norm and dominate the optimization process.
However, in
Figure~\ref{fig3:sub1}~\&~\ref{fig3:sub2},
when training a slimmable network
with widths $[1.0,0.5,0.25]$,
$\lVert \nabla_{\theta_{0.25}} \mathcal{L}_{\theta} \rVert_2$ becomes close or larger than
 $\lVert \nabla_{\theta_{1.0 \backslash 0.25}} \mathcal{L}_{\theta} \rVert_2$
despite having much fewer parameters.

Gradient magnitude imbalance is more pronounced in self-supervised cases.
In the case of supervised learning shown in Figure~\ref{fig3:sub1},
$\lVert \nabla_{\theta_{1.0 \backslash 0.25}}
\mathcal{L}_{\theta} \rVert_2$
is is initially close to
$\lVert \nabla_{\theta_{0.25}} \mathcal{L}_{\theta}\rVert_2$,
but the former increases as training progresses.
By contrast, for vanilla SlimCLR-MoCov2 in Figure~\ref{fig3:sub2},
$\lVert \nabla_{\theta_{1.0 \backslash 0.25}} \mathcal{L}_{\theta} \rVert_2$
is smaller than the other most time.
A conjecture is that the pretext task ---
instance discrimination is
more challenging than supervised classification.
Consequently, small networks with limited capacity face difficulty in converging, leading to large losses and dominant gradients.

\subsubsection{Gradient direction divergence}

In addition to the imbalance in gradient magnitudes,
the gradient directions of different weight-sharing networks can also conflict with each other.
These conflicts lead to disordered gradient directions of the full model, which we refer to as gradient direction divergence.

\begin{figure*}[!h]
	\centering
	\begin{subfigure}{.305\linewidth}
		\centering
		\includegraphics[width=\linewidth]{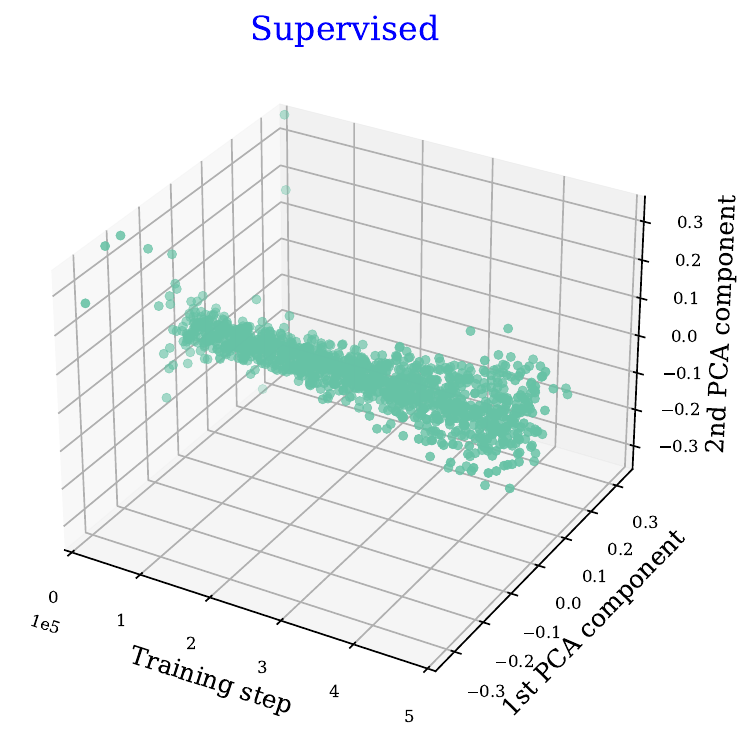}
  		\caption{supervised R-50$_{[1.0,0.5,0.25]}$}
		\label{fig-gd:sub11}
	\end{subfigure}%
	\begin{subfigure}{.34\linewidth}
		\centering
		\includegraphics[width=\linewidth]{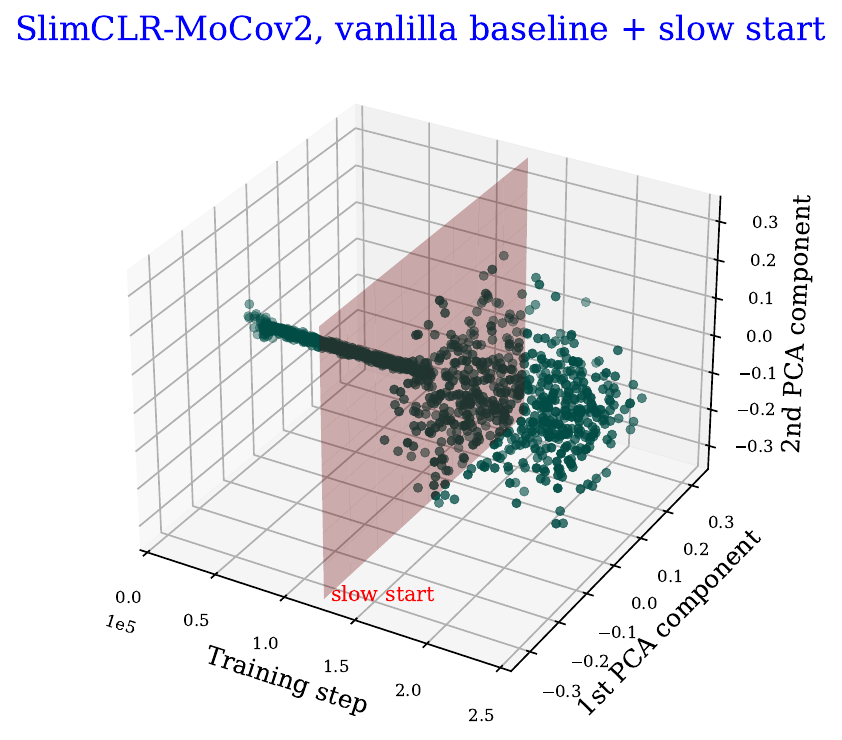}
  		\caption{SlimCLR-MoCov2}
		\label{fig-gd:sub12}
	\end{subfigure}%
	\begin{subfigure}{.33\linewidth}
		\centering
		\includegraphics[width=\linewidth]{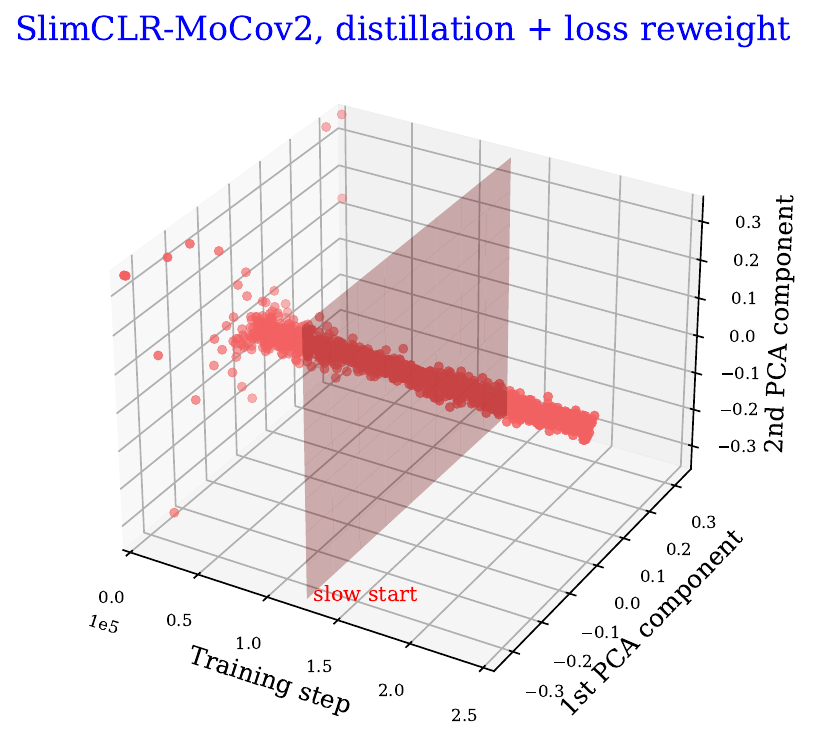}
  		\caption{SlimCLR-MoCov2}
		\label{fig-gd:sub13}
	\end{subfigure}%
	\caption{Gradient direction divergence.
    The principal gradient directions of the last linear layer are shown. 
    }
	\label{fig-grad-diverge}
\end{figure*}

We visualize
the principal gradient directions in Figure~\ref{fig-grad-diverge}.
Specifically, we collect the gradients with respect to
parameters in the last linear layer during training;
after training,
we perform PCA on these gradients and calculate their projections on the first two principal components~\citep{2018_vis}.
In Figure~\ref{fig-gd:sub12},
before the slow start point, we only train the full model.
In this case, the gradient directions are stable and consistent during training.
By contrast, after the slow start point,
gradient directions become disordered due to conflicts in weight-sharing networks.
We also show the gradient directions when training a slimmable
network in a supervised case in Figure~\ref{fig-gd:sub11}.
In the supervised case, gradient direction divergence is also inevitable.
Nevertheless, the gradient direction divergence
in the supervised case is less significant compared
to the self-supervised case in Figure~\ref{fig-gd:sub12}.
In supervised cases, all networks have consistent global
supervision during training.
By contrast, in self-supervised cases, the negative samples in Eq.~\eqref{infonce} are always changing during training, and the optimization goal is not consistent throughout the training process. Additionally, since the predictions of weight-sharing networks may vary significantly from each other,
this can result in different similarity scores in Eq.~\eqref{infonce} and divergent gradient directions. These factors exacerbate the phenomenon of gradient direction divergence in self-supervised cases.

\subsubsection{Solutions to network interference}

To address the issue of gradient magnitude imbalance, it is 
necessary for the majority of parameters to dominate the 
optimization process, namely, the ratios of gradient norms
in Figure~\ref{fig3-grad} should be large.
To alleviate the problem of gradient 
direction divergence, constraints can be placed on the similarity 
scores in Eq.~\eqref{infonce} to prevent predictions from varying 
significantly from each other.
To achieve these goals, we develop 
three simple yet effective pre-training techniques: \emph{slow 
start}, \emph{online distillation}, and \emph{loss reweighting}. 
Furthermore, we provide theoretical evidence demonstrating that a 
weight-sharing linear layer is suboptimal during linear 
evaluation, and a \emph{switchable linear probe layer} is a 
better alternative.

\paragraph{\textbf{Slow start}}
At the start of training,
the model and contrastive similarity scores in Eq.~\eqref{infonce}
update quickly, resulting in an unstable optimization procedure. 
To prevent interference between networks making the situation 
harder, we employ a \emph{slow start} technique in which we only 
train the full model, updating $\theta_{1.0}$ by 
$\nabla_{\theta_{1.0}}\mathcal{L}_{\theta}$,
for the first $S$ epochs.
As shown in Figure~\ref{fig3:sub3}, the ratios of gradient 
norms are large prior to the $S$-th epoch, but dramatically 
decrease after the slow start point.
During the first $S$ epochs, the full model 
can learn knowledge from the data without disturbances, and 
sub-networks can inherit this knowledge via the shared 
parameters and  begin training with a better initialization. 
Similar approaches are also adopted in some one-shot NAS 
methods~\citep{2020_ofa,2020_bignas}.

\paragraph{\textbf{Online distillation}}
The full network has the highest capacity to acquire knowledge 
from the data,
and its predictions can provide guidance to all 
sub-networks in resolving gradient direction conflicts among 
weight-sharing networks.
Namely, sub-networks should learn from the full network.
Following~\cite{2019_unslim},
we minimize the KL divergence between the estimated probabilities of sub-networks and the full network:
\begin{align} \label{eq_online}
p_{w_i} &= \frac{ \me^{\overline{z}_{\xi_{w_1}}\cdot 
 \frac{\overline{z}_{\theta_{w_i}} }{\tau_2}}
 }{\me^{\overline{z}_{\xi_{w_1}}\cdot
 \frac{\overline{z}_{\theta_{w_i}}}{\tau_2}}
  + \sum_{z^-}\me^{z^-\cdot
 \frac{\overline{z}_{\theta_{w_i}}}{\tau_2}}}, \\
	\mathcal{L}_{p_{w_i}} &= - p_{w_1} \log p_{w_i},\quad
    w_i \in \{w_2,\ldots,w_n\}.
\end{align}
Here $\tau_2$ is a temperature coefficient of distillation.
The results in Figure~\ref{fig3:sub4} demonstrate that online distillation helps alleviate gradient magnitude imbalance, with the ratios $\frac{\lVert \nabla_{\theta_{1.0 \backslash 0.25}} \mathcal{L}_{\theta} \rVert_2}{ \lVert \nabla_{\theta_{0.25}} \mathcal{L}_{\theta} \rVert_2}$ becoming larger than $1.0$. This indicates that online distillation assists the majority of parameters in dominating the optimization process.

\paragraph{\textbf{Loss reweighting}}
Another approach to address interference among
weight-sharing networks is to assign large confidence to networks with large widths, allowing the strongest network to take control.
The weight for the loss of the network with width $w_i$ is:
\begin{align}
	\lambda_i = 1.0 + \1\{w_i=w_1\} \times \sum_{j=2}^{n}w_j,
\end{align}
where $\1\{\cdot\}$ equals to $1$ if the inner condition is true, 0 otherwise.
In Figure~\ref{fig3:sub5}, both ratios become large,
and $\frac{\lVert \nabla_{\theta_{1.0 \backslash 0.25}} \mathcal{L}_{\theta} \rVert_2}{ \lVert \nabla_{\theta_{0.25}} \mathcal{L}_{\theta} \rVert_2}$
is significantly larger than $1.0$.
In Figure~\ref{fig-gd:sub13},
using online distillation and loss reweighting leads to much more stable gradient directions compared to
compared to Figure~\ref{fig-gd:sub12}.

Considering online distillation and loss reweighting,
the overall pre-training objective is:
\begin{align}
	\mathcal{L}_{\theta} = \lambda_1\mathcal{L}_{\theta_{w_1}} +
	\sum_{i=2}^{n}\lambda_i
	\frac{\mathcal{L}_{\theta_{w_i}}
	+ \mathcal{L}_{p_{w_i}}}{2}.
\end{align}

\noindent\textbf{switchable linear probe layer~}
As we demonstrate theoretically in Appendix~\ref{app1}, given the 
features extracted by a slimmable network that is pre-trained via 
contrastive self-supervised learning, a single slimmable linear probe 
layer cannot achieve several complex mappings from different 
representations to the same object classes simultaneously. The 
failure is because the learned representations in 
Figure~\ref{fig2_framework} do not meet the requirement discussed in 
Appendix~\ref{app1}. In this case, we propose to use a switchable linear probe layer during linear evaluation,
where each network in the slimmable network 
has its own linear probe layer. This allows 
each network to learn its unique mapping from its representation to 
the object classes.
\section{Experiments}
\begin{table*}[!t]
    \centering
	\scalebox{0.88}{
		\begin{tabular}{l|lc|cc|crr}
			\toprule 
			Method
			& Backbone   & Teacher
			& Top-1      & Top-5
			& Epochs
			& \#Params   & \#FLOPs \\
			\midrule

			\multirow{7}{*}{Supervised}
			& R-50
			& \multirow{3}{*}{\xmark}
			& 76.6      & 93.2
			& 100
			& 25.6 M    & 4.1 G  \\
            ~
			& R-34
			&
			& 75.0
			& -
			& -
			& 21.8~M
			& 3.7~G \\
			~
			& R-18
			&
			& 72.1
			& -
			& -
			& 11.9~M
			& 1.8~G \\
            \cdashline{2-8}[2pt/2pt]
			
			& R-50$_{1.0}$
			& \multirow{4}{*}{\xmark}
			& 76.0{\color{cyan}$_{(\bm{0.6}\downarrow)}$}
			& 92.9
			& \multirow{4}{*}{100}
			& \multirow{1}{*}{25.6 M}
			& \multirow{1}{*}{4.1 G}\\

			~
			& R-50$_{0.75}$
			& 
			& 74.9
			& 92.3
			& 
			& \multirow{1}{*}{14.7 M}
			& \multirow{1}{*}{2.3~G}  \\

			~
			& R-50$_{0.5}$
			&
			& 72.2
			& 90.8
			& 
			& 6.9~M
			& 1.1~G \\

			~
			& R-50$_{0.25}$
			&
			& 64.4
			& 86.0
			& 
			& 2.0~M
			& 278~M \\
            
			\midrule
            
			\multirow{5}{*}{\shortstack{Baseline \\ (individual networks \\ trained with MoCov2)}}
			& R-50
			& \multirow{1}{*}{\xmark}
			& 67.5
			& -
			& \multirow{1}{*}{200}
			& 25.6 M
			& 4.1 G  \\

            \cdashline{2-8}[2pt/2pt]

			& R-50$_{1.0}$
			& \multirow{4}{*}{\xmark}
			& 67.2
			& 87.8
			& \multirow{4}{*}{200}
			& \multirow{1}{*}{25.6 M}
			& \multirow{1}{*}{4.1 G}\\
			
			~
			& R-50$_{0.75}$
			& 
			& 64.3
			& 85.8
			& 
			& \multirow{1}{*}{14.7 M}
			& \multirow{1}{*}{2.3~G}  \\
			
			~
			& R-50$_{0.5}$
			&
			& 58.9
			& 82.2
			& 
			& 6.9~M
			& 1.1~G \\
			
			~
			& R-50$_{0.25}$
			&
			& 47.9
			& 72.8
			& 
			& 2.0~M
			& 278~M \\
			\midrule

			MoCov2~\citeyearpar[preprint]{mocov2}
			& R-50
			& \multirow{5}{*}{\xmark}
			& 71.1
			& -
			& 800
			& \multirow{5}{*}{25.6 M}
			& \multirow{5}{*}{4.1 G} \\

			MoCov3~\citeyearpar[ICCV]{2021_mocov3}
			& R-50
			& 
			& 72.8
			& -
			& 300
			&
			& \\

			SlimCLR-MoCov2
			& R-50$_{1.0}$
			& 
			& 67.4{\color{cyan}$_{(\bm{0.1}\downarrow)}$}
			& 87.9
			& 200
			& 
			& \\
			
			SlimCLR-MoCov2
			& R-50$_{1.0}$
			& 
			& 70.4{\color{cyan}$_{(\bm{0.7}\downarrow)}$}
			& 89.6
			& 800
			& 
			& \\

			SlimCLR-MoCov3
			& R-50$_{1.0}$
			& 
			& 72.3{\color{cyan}$_{(\bm{0.5}\downarrow)}$}
			& 90.8
			& 300
			& 
			& \\

			\midrule

			SEED~\citeyearpar[ICLR]{seed}
			& R-34
			& R-50~(67.4)
			& 58.5
			& 82.6 
			& 200
			& \multirow{6}{*}{21.8 M}
			& \multirow{6}{*}{3.7~G}  \\

			DisCo~\citeyearpar[ECCV]{disco}
			& R-34
			& R-50~(67.4)
			& 62.5
			& 85.4 
			& 200
			& 
			& \\

			BINGO~\citeyearpar[ICLR]{2022_bingo}
			& R-34
			& R-50~(67.4)
			& 63.5
			& 85.7 
			& 200
			& 
			& \\

			SEED~\citeyearpar[ICLR]{seed}
			& R-34
			& R-50$\times$2~(77.3)
			& 65.7
			& 86.8 
			& 800
			& ~
			& ~  \\

			DisCo~\citeyearpar[ECCV]{disco}
			& R-34
			& R-50$\times$2~(77.3)
			& 67.6
			& 88.6 
			& 200
			& 
			& \\

			BINGO~\citeyearpar[ICLR]{2022_bingo}
			& R-34
			& R-50$\times$2~(77.3)
			& 68.9
			& 88.9
			& 200
			& 
			& \\

            \cdashline{2-8}[3pt/2pt]
			SlimCLR-MoCov2
			& R-50$_{0.75}$
			& \multirow{3}{*}{\xmark}
			& 65.5
			& 87.0
			& 200
			& \multirow{3}{*}{\color{blue}{14.7 M}}
			& \multirow{3}{*}{\color{blue}{2.3 G}}\\

			SlimCLR-MoCov2
			& R-50$_{0.75}$
			& 
			& 68.8
			& 88.8
			& 800
			& 
			&  \\

			SlimCLR-MoCov3
			& R-50$_{0.75}$
			& 
			& \color{blue}{\textbf{69.7}}
			& \color{blue}{\textbf{89.4}}
			& 300
			& 
			&  \\
			\midrule

			CompRess~\citeyearpar[NeurIPS]{compress}
			& R-18
			& R-50~(71.1)
			& 62.6
			& - 
			& 130
			& \multirow{7}{*}{11.9 M}
			& \multirow{7}{*}{1.8~G}  \\

			SEED~\citeyearpar[ICLR]{seed}
			& R-18
			& R-50$\times$2~(77.3)
			& 63.0
			& 84.9 
			& 800
			& ~
			& ~  \\

			DisCo~\citeyearpar[ECCV]{disco}
			& R-18
			& R-50$\times$2~(77.3)
			& 65.2
			& 86.8 
			& 200
			& 
			& \\

			BINGO~\citeyearpar[ICLR]{2022_bingo}
			& R-18
			& R-50$\times$2~(77.3)
			& 65.5
			& 87.0
			& 200
			& 
			& \\

			SEED~\citeyearpar[ICLR]{seed}
			& R-18
			& R-152~(74.1)
			& 59.5
			& 65.5 
			& 200
			& ~
			& ~  \\

			DisCo~\citeyearpar[ECCV]{disco}
			& R-18
			& R-152~(74.1)
			& 65.5
			& 86.7 
			& 200
			& 
			& \\

			BINGO~\citeyearpar[ICLR]{2022_bingo}
			& R-18
			& R-152~(74.1)
			& 65.9
			& 87.1 
			& 200
			& 
			& \\
            \cdashline{2-8}[2pt/2pt]

			Shi~\textit{et al.}~\citeyearpar[AAAI]{shi2022efficacy}
			& R-18
			& \multirow{4}{*}{\xmark}
			& 55.7 & - 
			& 800
			& 11.9~M
			& 1.8~G\\		
	
			SlimCLR-MoCov2
			& R-50$_{0.5}$
			& 
			& 62.5
			& 84.8
			& 200
			&\multirow{3}{*}{\color{blue}{6.9~M}}
			&\multirow{3}{*}{\color{blue}{1.1~G}}\\
			
			SlimCLR-MoCov2
			& R-50$_{0.5}$
			& 
			& {65.6}
			& {87.2}
			& 800
			& 
			&  \\

			SlimCLR-MoCov3
			& R-50$_{0.5}$
			& 
			& \color{blue}{\textbf{67.6}}
			& \color{blue}{\textbf{88.2}}
			& 300
			& 
			&  \\
			\midrule

			Shi~\textit{et al.}~\citeyearpar[AAAI]{shi2022efficacy}
			& Eff-B0\textsuperscript{\S}
			& \multirow{4}{*}{\xmark}
			& 51.2 & - 
			& 200
			& 5.3~M
			& 390~M\\	

			SlimCLR-MoCov2
			& R-50$_{0.25}$
			& 
			& 55.1
			& 79.5
			& 200
			&\multirow{3}{*}{\color{blue}{2.0~M}}
			&\multirow{3}{*}{\color{blue}{278~M}}\\

			SlimCLR-MoCov2
			& R-50$_{0.25}$
			& 
			& {57.6}
			& {81.5}
			& 800
			& 
			&  \\

			SlimCLR-MoCov3
			& R-50$_{0.25}$
			& ~
			& \color{blue}{\textbf{62.4}}
			& \color{blue}{\textbf{84.4}}
			& 300
			& 
			&  \\
			\bottomrule
		\end{tabular}
}
	\caption{Linear evaluation results of SlimCLR with ResNet-50$_{[1.0,0.75,0.5,0.25]}$
    on ImageNet.
    Through only one-time pre-training,
    SlimCLR produces multiple different small models without extra large teacher models.
Plus, SlimCLR outperforms previous methods with fewer parameters and FLOPs.
    The {\color{cyan}performance degradation} when training
    a slimmable network is shown in cyan.
    \textsuperscript{\S}Eff-B0 is short for EfficientNet-B0~\citep{tan2019efficientnet}.}
	\label{tab:imagenet_res}
\end{table*}

\subsection{Experimental details}
\textbf{Datatest~}
We train SlimCLR on ImageNet~\citep{ImageNet}, which contains 1.28M training and 50K validation images. During pre-training, we use training images
without labels.

\noindent\textbf{Pre-training of SlimCLR-MoCov2~}
By default, we use a total batch size 1024, an initial learning rate 0.2,
and weight decay $1\times 10^{-4}$.
We adopt the SGD optimizer with a momentum 0.9.
A linear warm-up and cosine decay 
policy~\citep{2017_PriyaGoyal,2018_bags_of_tricks} for learning rate is applied,
and the warm-up epoch is 10.
The temperatures are $\tau_1=0.2$ for InfoNCE  and $\tau_2=5.0$
for online distillation. 
Without special mentions, other settings including
data augmentations, queue size~(65,536), and 
feature dimension~(128) are the same as
the counterparts of MoCov2~\citep{mocov2}.
The slow start epoch $S$ of sub-networks
is set to be half of the number of total epochs.

\noindent\textbf{Pre-training of SlimCLR-MoCov3~}
We use a total batch size 1024,
an initial learning rate 1.2,
and weight decay $1\times 10^{-6}$.
We adopt the LARS~\citep{you2017large} optimizer and
a cosine learning rate policy with warm-up epoch 10.
The temperatures are $\tau_1=1.0$ and $\tau_2=1.0$.
The slow start epoch $S$ is half of the total epochs.
One different thing is that we increase the initial
learning rate to 3.2 after $S$ epochs.
Pre-training is all done with
mixed precision~\citep{2018_AMP}.

\noindent\textbf{Linear evaluation~}
Following the general linear evaluation protocol
~\citep{2020_simclr,2020_moco},
we add new linear layers on the backbone and freeze the backbone during evaluation.
We also apply online distillation
with a temperature $\tau_2=1.0$ when training these linear layers.
For evaluation of SlimCLR-MoCov2,
we use a total 
batch size 1024, epochs 100,
and an initial learning rate 60, which is decayed by
10 at 60 and 80 epochs.
For evaluation of SlimCLR-MoCov3,
we use a total batch size 1024, epochs 90,
and an initial learning rate 0.4 with cosine decay
policy.

\begin{table*}[!t]
	\centering
	\begin{subtable}{\linewidth}
		\centering
		\scalebox{0.88}{
			\begin{tabular}{l|lr|ccc|ccc}
				\toprule
				pre-train
				& backbone
                    & \#Params
				& \apbbox{}
				& \apbbox{50} 
				& \apbbox{75}
				& \apmask{}
				& \apmask{50}
				& \apmask{75} 
				\\
				\midrule

	  \multirow{4}{*}{supervised~\citep{2019_slim}} 
				& R-50$_{1.0}$
                    & 25.6~M
				& 37.4 & 59.6 & 40.5
				& 34.9 & 56.5 & 37.3    \\ 
				
				& R-50$_{0.75}$
                    & 14.7~M
				& 36.7 & 58.7 & 39.3
				& 34.3 & 55.4 & 36.1    \\
				
				& R-50$_{0.5}$
                    & 6.9~M
				& 34.7 & 56.3 & 36.8
				& 32.6 & 53.1 & 34.1    \\ 
				
				& R-50$_{0.25}$
                    & 2.0~M
				& 30.2 & 50.3 & 31.5
				& 28.6 & 47.5 & 29.9    \\ 
                \midrule
				MoCo~\citep{2020_moco}
				&  R-50
                    & 25.6~M
				& 38.5 & 58.9 & 42.0
				& 35.1 & 55.9 & 37.7    \\  
				SEED~\citep{seed}
				& R-34
                    & 21.8~M
				& 38.4 & 57.0 & 41.0
				& 33.3 & 53.6 & 35.4    \\
				BINGO~\citep{2022_bingo}
				& R-18
                    & 11.9~M
				& 32.0 & 51.0 & 34.7
				& 29.6 & 48.2 & 31.5    \\
                \midrule
				\multirow{4}{*}{SlimCLR-MoCov2} 
				& R-50$_{1.0}$
                    & 25.6~M
				& \textbf{38.6} & \textbf{60.1} & \textbf{42.0}
				& \textbf{35.7} & \textbf{57.2} & \textbf{38.0}    \\ 
				
				& R-50$_{0.75}$
                    & 14.7~M
				& 37.7 & \textbf{59.3} & 40.9
				& \textbf{34.9} & \textbf{56.3} & \textbf{37.4}    \\

				& R-50$_{0.5}$
                    & 6.9~M
				& \textbf{35.8} & \textbf{56.9} & \textbf{38.6}
				& \textbf{33.2} & \textbf{54.2} & \textbf{35.3}    \\

				& R-50$_{0.25}$
                    & 2.0~M
				& \textbf{31.1} & \textbf{51.0} & \textbf{33.3}
				& \textbf{29.1} & \textbf{48.1} & \textbf{30.9}    \\ 
				\bottomrule
			\end{tabular}
			}
		\label{tab3:sub1}
	\end{subtable}
	\caption{Transfer learning results of SlimCLR pre-trained models on COCO \texttt{val2017} set.
	Bounding-box AP (\apbbox{}) for \textbf{object detection} and mask AP (\apmask{}) for \textbf{instance segmentation}. The parameters of backbones during pre-training are also presented.
	 \label{table-coco}
	}
\vspace{-0.3cm}
\end{table*}


\subsection{Linear evaluation on ImageNet}
The results of SlimCLR on ImageNet are presented in 
Table~\ref{tab:imagenet_res}. Despite our efforts to 
mitigate the interference of weight-sharing networks, as 
described in Section~\ref{sec-grad}, slimmable training 
unavoidably causes a drop in performance for the full 
model. Moreover, when training for more epochs, the 
performance degradation becomes more pronounced.
However, it is important to note that such degradation is 
not unique in the self-supervised case, and it also occurs 
in the supervised case.
Despite this drop in performance, slimmable training has 
significant advantages, as we will discuss below.

The results of SlimCLR on ImageNet show that it can help sub-networks achieve significant performance improvements compared to MoCov2 with individual networks. Specifically, when pre-training for 200 epochs, SlimCLR-MoCov2 achieves 3.5\% and 6.6\% improvements in performance for ResNet-50$_{0.5}$ and ResNet-50$_{0.25}$, respectively. This indicates that sub-networks can inherit knowledge from the full model via parameter sharing to enhance their generalization ability. Moreover, using a more powerful contrastive learning framework, such as SlimCLR-MoCov3, can further boost the performance of sub-networks.

Compared to previous methods aimed at distilling knowledge from large teacher models, sub-networks of ResNet-50$_{[1.0, 0.75, 0.5, 0.25]}$ achieve superior performance with fewer parameters and FLOPs. Notably, SlimCLR also helps smaller models approach the performance of their supervised counterparts. In addition, SlimCLR obviates the need for any additional training of large teacher models, and all networks in SlimCLR are jointly trained. By training only once, we obtain different models with varying computational costs that are suitable for different devices. These results demonstrate the superiority of adopting slimmable networks for contrastive learning, and highlight the potential of SlimCLR in various applications.

\begin{table*}[!t]
	\centering
	\begin{subtable}{0.32\linewidth}
		\centering
		\scalebox{0.8}{
			\begin{tabular}{l|cccc}
				\toprule
				\multirow{2}{*}{Model}
				& \multicolumn{2}{c}{slimmable}
				& \multicolumn{2}{c}{switchable}   \\
				
				& Top-1 & Top-5
				& Top-1 & Top-5 \\
				\midrule
				R-50$_{1.0}$  &  64.8 & 86.1 
				& \textbf{65.6} & \textbf{86.8} \\ 
				
				R-50$_{0.75}$ &  63.4 & 85.3 
				& \textbf{64.3} & \textbf{86.0} \\
				
				R-50$_{0.5}$  &  59.6 & 82.9 
				& \textbf{61.3} & \textbf{84.1} \\
				
				R-50$_{0.25}$ &  53.0 & 77.8
				& \textbf{54.5} & \textbf{79.1} \\
				\bottomrule
			\end{tabular}
			}
		\caption{switchable linear probe layer}
		\label{tab2:sub1}
	\end{subtable}
	\begin{subtable}{0.32\linewidth}
	\centering
	\scalebox{0.8}{
		\begin{tabular}{l|cccc}
			\toprule
			\multirow{2}{*}{Model}
			& \multicolumn{2}{c}{$S=0$}
			& \multicolumn{2}{c}{$S=100$}   \\
			& Top-1 & Top-5
			& Top-1 & Top-5 \\
			\midrule
			R-50$_{1.0}$  &  65.6 & 86.8
			& \textbf{66.7} & \textbf{87.5} \\ 
			
			R-50$_{0.75}$ &  64.3 & 86.0 
			& \textbf{65.3} & \textbf{86.4} \\
			
			R-50$_{0.5}$  &  61.3 & 84.1
			& \textbf{62.5} & \textbf{84.3} \\
			
			R-50$_{0.25}$ &  54.5 & 79.1
			& \textbf{54.9} & \textbf{79.5} \\
			\bottomrule
		\end{tabular}
	}
	\caption{slow start epoch $S$, 200 epochs}
	\label{tab2:sub2}
	\end{subtable}
	\begin{subtable}{0.32\linewidth}
	\centering
	\scalebox{0.8}{
		\begin{tabular}{l|cccc}
			\toprule
			\multirow{2}{*}{Model}
			& MSE
			& ATKD
			& DKD   
			& KD \\
			& \multicolumn{4}{c}{Top-1} \\
			\midrule
			R-50$_{1.0}$  &  66.9 & 66.4
			& {66.8} & \textbf{67.0} \\ 
			
			R-50$_{0.75}$ &  {65.2} & 65.0 
			& {65.1} & \textbf{65.3} \\
			
			R-50$_{0.5}$  &  62.4 & 62.2
			& {62.3} & \textbf{62.6} \\

			R-50$_{0.25}$ &  \textbf{55.2} & 54.7
			& {54.9} & 54.9 \\
			\bottomrule
		\end{tabular}
	}
	\caption{online distillation loss choice}
	\label{tab2:sub3}
	\end{subtable}
	
	\begin{subtable}{0.32\linewidth}
	\centering
	\scalebox{0.8}{
		\begin{tabular}{l|cccc}
			\toprule
			\multirow{2}{*}{Model}
			& $3.0$
			& $4.0$
			& $5.0$   
			& $6.0$ \\
			& \multicolumn{4}{c}{Top-1} \\
			\midrule
			R-50$_{1.0}$  &  \textbf{67.0} & \textbf{67.0}
			& \textbf{67.0} &  66.7 \\ 
			
			R-50$_{0.75}$ &  {65.2} & \textbf{65.3 }
			& \textbf{65.3} & 65.2 \\
			
			R-50$_{0.5}$  &  62.4 & \textbf{62.6}
			& {\textbf{62.6}} & 62.4 \\
			
			R-50$_{0.25}$ &  \textbf{55.0} & 54.8
			& {54.9} & \textbf{55.0} \\
			\bottomrule
		\end{tabular}
	}
	\caption{online distillation temperature}
	\label{tab2:sub4}
	\end{subtable}
	\begin{subtable}{0.32\linewidth}
	\centering
	\scalebox{0.8}{
		\begin{tabular}{l|cccc}
			\toprule
			\multirow{2}{*}{Model}
			& (1)
			& (2)
			& (3)   
			& (4) \\
			& \multicolumn{4}{c}{Top-1} \\
			\midrule
			R-50$_{1.0}$  &  67.4 & 67.3
			& \textbf{67.5} &  \textbf{67.5} \\ 
			
			R-50$_{0.75}$ &  {65.5} & 65.7
			& \textbf{65.9} & 65.8 \\
			
			R-50$_{0.5}$  &  62.5 & 62.2
			& \textbf{62.6}  & 62.4 \\
			
			R-50$_{0.25}$ &  \textbf{55.1} & 54.5 
			& 54.4 & 54.5 \\
			\bottomrule
		\end{tabular}
	}
	\caption{loss reweighting}
	\label{tab2:sub5}
	\end{subtable}
	\begin{subtable}{0.32\linewidth}
	\centering
	\scalebox{0.8}{
		\begin{tabular}{l|cccc}
			\toprule
			\multirow{2}{*}{Model}
			& $200$
			& $300$
			& $400$
			& $500$ \\

			& \multicolumn{4}{c}{Top-1} \\
			\midrule
			R-50$_{1.0}$  &  70.2 & 70.0
			& \textbf{70.3} & 70.1 \\ 
			
			R-50$_{0.75}$ &  68.3 & 68.6 
			& \textbf{68.8} & 68.4 \\
			
			R-50$_{0.5}$  &  \textbf{65.7} & 65.6
			& 65.6 & 65.3 \\
			
			R-50$_{0.25}$ &  \textbf{57.8} & 57.5
			& 57.6 & 57.3 \\
			\bottomrule
		\end{tabular}
	}
	\caption{slow start epoch $S$, 800 epochs}
	\label{tab2:sub6}
	\end{subtable}
	\caption{Ablation experiments with SlimCLR-MoCov2 on ImageNet.
	The experiment in a former table serves
	as a baseline for the consequent table.
	}
\end{table*}

\subsection{Transfer learning}
In this section, we assess the transfer learning 
ability of SlimCLR on object detection and 
instance segmentation, using the Mask
R-CNN~\citep{2017_maskrcnn} with
FPN~\citep{2017_fpn} architectures as the supervised 
slimmable network~\citep{2019_slim}.
We fine-tune all parameters, including 
batch normalization~\citep{2015_bn}, end-to-end 
on the COCO~2017 dataset~\citep{2014_coco} using 
a default $1\times$ training schedule from 
MMDetection~\citep{mmdetection}.
During training, we apply synchronized batch normalization across different GPUs.
The backbone is a ResNet-50$_{[1.0,0.75,0.5,0.25]}$ pre-trained via SlimCLR-MoCov2 for 800 epochs.
Our results, presented in Table~\ref{table-coco}, demonstrate that SlimCLR-MoCov2 outperforms the supervised baseline in terms of transfer learning ability. Notably, by training only once, SlimCLR produces multiple networks that surpass previous methods pre-trained with large teachers while requiring fewer parameters. Particularly, SlimCLR outperforms previous distillation-based pre-training methods significantly in the instance segmentation task.
These findings demonstrate the effectiveness of pre-training with SlimCLR.

\subsection{Discussion} \label{sec_discussion}
In this section, we will discuss the influences of different components in SlimCLR.

\noindent\textbf{switchable linear probe layer~} 
Table~\ref{tab2:sub1} demonstrates the notable 
impact of the switchable linear probe layer compared 
to a slimmable linear probe layer during linear evaluation.
The introduction 
of a switchable linear probe layer results in 
significant improvements in accuracy.
When only one slimmable layer is used,
the interference 
between weight-sharing linear layers is 
unavoidable as discussed in Appendix~\ref{app1}.
The learned representations of pre-trained models
do not satisfy the requirements in Appendix~\ref{app1}.

\begin{table}[t]
    \centering
    \resizebox{0.98\linewidth}{!}{
    \begin{tabular}{l|l|c|c}
         Method & Backbone & \#V100 & Time~(h)  \\
         \midrule
         SwAV~(\citeauthor{caron2020unsupervised})
         & R-50$\times$2 & 128 & 36.0 \\
         SwAV~(\citeauthor{caron2020unsupervised})
         & R-50 & 64 & 12.5 \\
          MoCov2~(\citeauthor{shi2022efficacy})
          & R-50 & 8 & 42.4
         \\
         MoCov2~(\citeauthor{shi2022efficacy})
         & R-18 & 8 & 40.9 \\
         \midrule
         SlimCLR-MoCov2 & R-50$_{[1.0]}$ & 8 & 19.1 \\
         SlimCLR-MoCov2 & R-50$_{[1.0,0.5]}$ & 8 & 23.6 \\
         SlimCLR-MoCov2 & R-50$_{[1.0,0.5,0.25]}$ & 8 & 26.6 \\
         SlimCLR-MoCov2 & R-50$_{[1.0,0.75,0.5,0.25]}$ & 8 & 33.4 \\
         \bottomrule
    \end{tabular}
    }
    \caption{The pre-training time for 200 epochs. \#V100 is the number of Tesla V100 GPUs.
    The slow start epoch $S=100$.
    We adopt NVIDIA DALI~\citep{dali} to accelerate the data augmentation pipeline.}
     \label{tab:training-time}
\end{table}

\noindent\textbf{slow start and training time~}
Table~\ref{tab2:sub2} presents experiments conducted 
with and without the slow start technique.
The use of slow start prevents interference between weight-sharing networks during the initial stages of training, allowing the system to quickly reach a stable point during optimization. 
Consequently, sub-networks start with better initialization and achieve better performance.
Furthermore, Table~\ref{tab2:sub6} provides ablations of the slow start epoch $S$ when training for a longer duration. Choosing $S$ to be half of the total epochs is a natural and appropriate choice.

Table~\ref{tab:training-time} illustrates the pre-training time of SlimCLR-MoCov2.
Pre-training a teacher model is expensive, especially when the model is large, \eg, ResNet-50$\times2$ and ResNet-152 in Table~\ref{tab:imagenet_res}.
Distillation methods also involve an additional knowledge transfer process, the cost of which is not negligible, as suggested by \cite{shi2022efficacy}.
Moreover, distillation methods once can only yield a single small model.
In contrast, by extending the pre-training time of MoCov2 (ResNet-50) by 14.3 hours, we obtain SlimCLR-MoCov2 with ResNet-50$_{[1.0,0.75,0.5,0.25]}$, which includes 4 various networks suitable for different computation scenarios.
The extra pre-training time can be further reduced if we use ResNet-50$_{[1.0,0.5.0.25]}$~(+7.5 hours) or ResNet-50$_{[1.0,0.5]}$~(+4.5 hours).
In a word, SlimCLR offers versatility and high computational efficiency.

\noindent\textbf{online distillation~}
We compared four different distillation losses, including the classical mean-square-error~(MSE) and KL divergence (KD), as well as two recent approaches: ATKD~\citep{guo2022reducing}
and DKD~\citep{zhao2022decoupled}.
ATKD reduces the difference in sharpness between distributions
of teacher and student to help the student better mimic the teacher. 
DKD decouples the classical knowledge distillation objective function
into target class and non-target class knowledge distillation to achieve
more effective and flexible distillation.
In Table~\ref{tab2:sub3}, these four distillation losses make trivial differences in our context.

Upon combining the outcomes of distillation with ResNet-50$_{[1.0,0.5,0.25]}$ in 
Figure~\ref{fig3:sub4}, we observe that the primary 
advantage of distillation lies in improving the 
performance of the full model, while the 
enhancements in sub-networks are relatively 
marginal. This contradicts the original objective of 
knowledge distillation, which aims to transfer 
knowledge from larger models to smaller ones and 
enhance their performance. One plausible explanation 
for this outcome is that the sub-networks in a 
slimmable network already inherit the knowledge from 
the full network through shared parameters, rendering logtis distillation less useful. In our context, 
the primary function of online distillation is to 
alleviate the interference among weight-sharing sub-networks, as demonstrated in Figure~\ref{fig3:sub4}~\&~\ref{fig-gd:sub13}.

We also test the influence of different temperatures in online distillation, \ie,
$\tau_2$ in Eq.~\eqref{eq_online}.
Following classical KD~\citep{2015_kd}, we choose $\tau_2 \in \{3.0, 4.0, 5.0, 6.0\}$.
The results are presented in Table~\ref{tab2:sub4}.
The choices of temperatures make trivial differences.
By contrast,
SEED~\citep{seed} uses a small temperature 0.01 for 
the teacher to get sharp distribution and a relatively large
temperature 0.2 for the student.
BINGO~\citep{2022_bingo} adopts a single temperature 0.2. These choices are different from ours, but we 
observed that SlimCLR is more robust to the choice of temperature.

~\\
\textbf{loss reweighting~}
We compared four loss reweighting manners in 
Table~\ref{tab2:sub5}.
They are {\small
\begin{align}
   &(1).~~ \lambda_i = 1.0 + \1\{w_i=w_1\} \times \sum_{j=2}^{n}w_j, \notag \\
   &(2).~~\lambda_i = 1.0 + \1\{w_i=w_1\} \times \max\{w_2,\ldots,w_n\}, \notag \\
   &(3).~~\lambda_i = n \times \frac{w_i}{\sum_{j=1}^{n}w_j}, \notag \\
   &(4).~~\lambda_i = n \times \frac{1.0 + \sum_{j=i}^{n}w_j}{\sum_{j=1}^{n} (1.0 + \sum_{k=j}^{n}w_k)}. \notag
\end{align}
}where $\1\{\cdot\}$ equals to $1$ if the inner 
condition is true, 0 otherwise.
The corresponding weights of 
networks with widths $[1.0,0.75,0.5,0.25]$
are $[2.5, 1.0, 1.0, 1.0]$,
$[1.75, 1.0, 1.0, 1.0]$,
$[1.6, 1.2, 0.8, 0.4]$, and
$[1.54, 1.08, 0.77, 0.62]$.
It is clear that a larger weight for the full model 
helps the system achieve better performance.
This demonstrates again that it is important for the 
full model to lead the training process.
The differences between the above four loss reweighting strategies are mainly 
reflected in the sub-networks with small sizes.
To ensure the performance of the smallest network,
we adopt the reweighting manner (1) in practice.

\begin{figure}[!t]
	\centering
	\begin{subfigure}{\linewidth}
		\centering
		\includegraphics[width=0.49\linewidth]{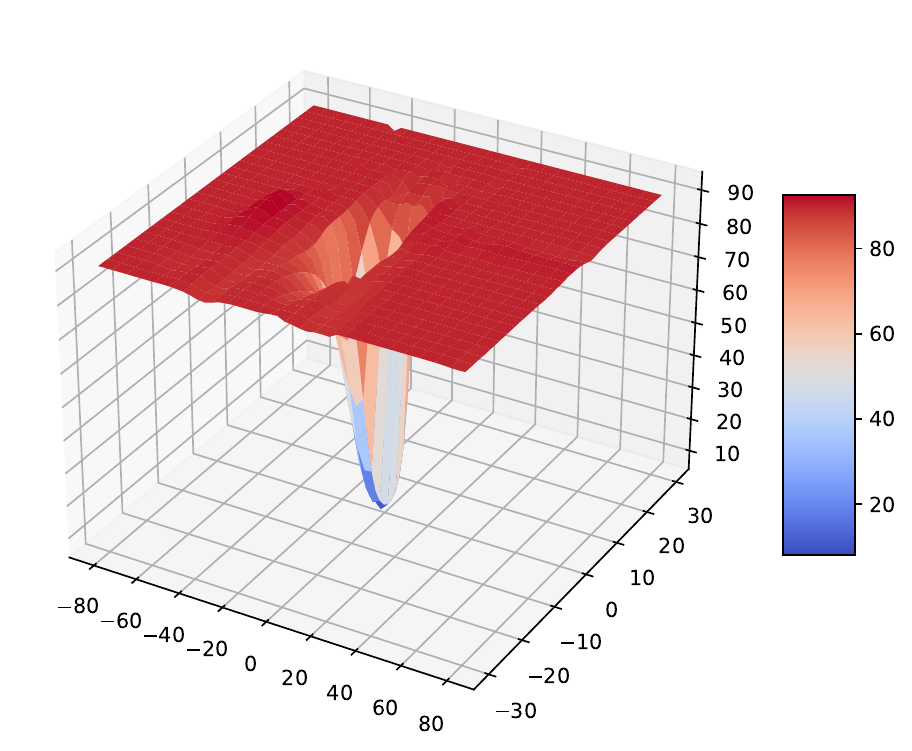}
		\includegraphics[width=0.49\linewidth]{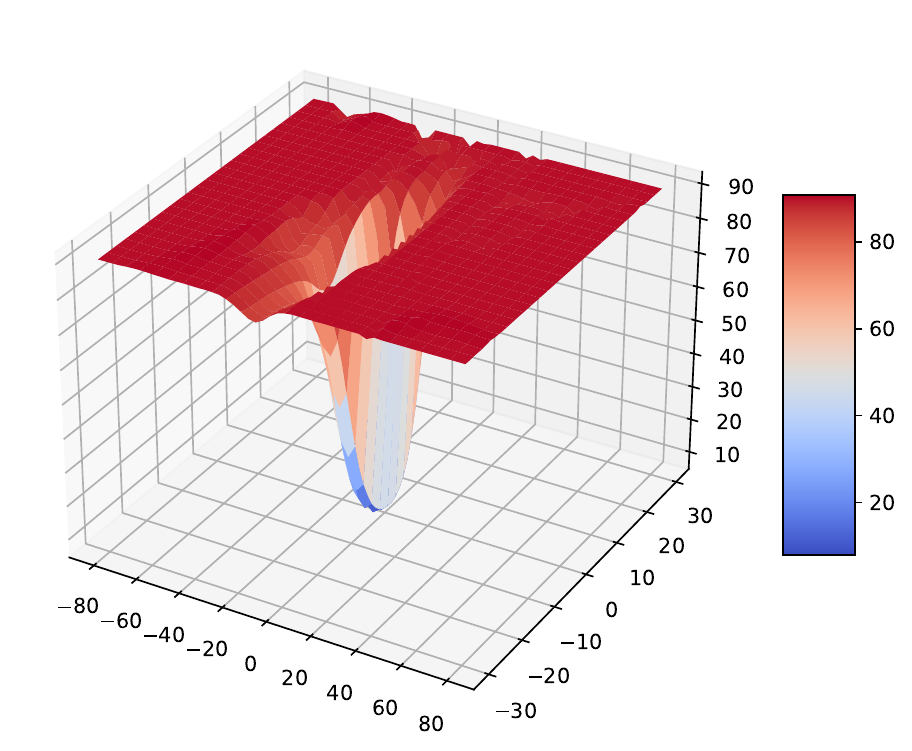}
		\caption{error surface (supervised). left: R-20$\times$4, 94.07\%;
		right: R-20$\times$4$_{[1.0,0.5]}$,
		93.58\%, 93.03\%.}
		\label{fig:vis1}
	\end{subfigure}
	\begin{subfigure}{\linewidth}
		\centering
		\includegraphics[width=0.48\linewidth]{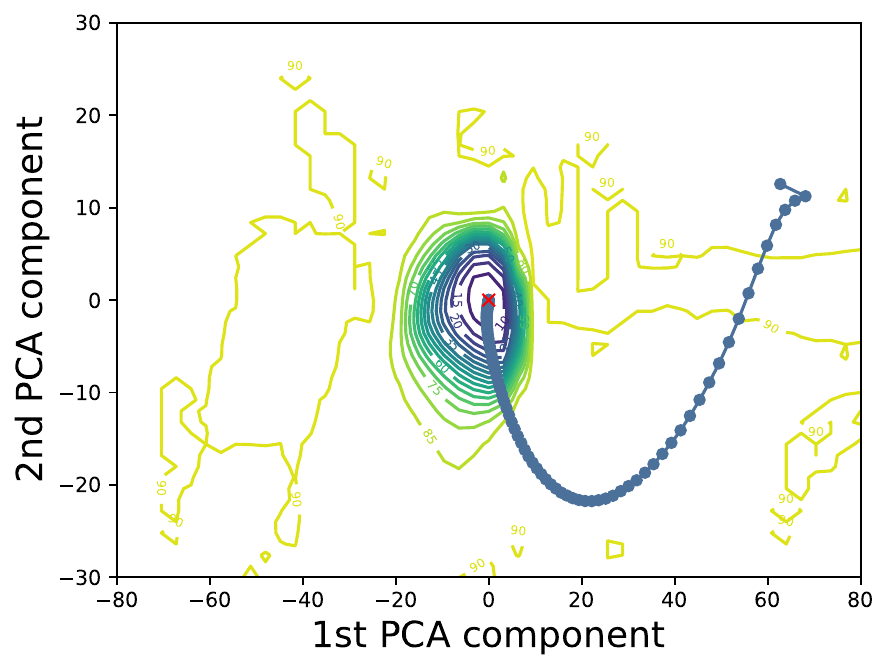}
		\includegraphics[width=0.48\linewidth]{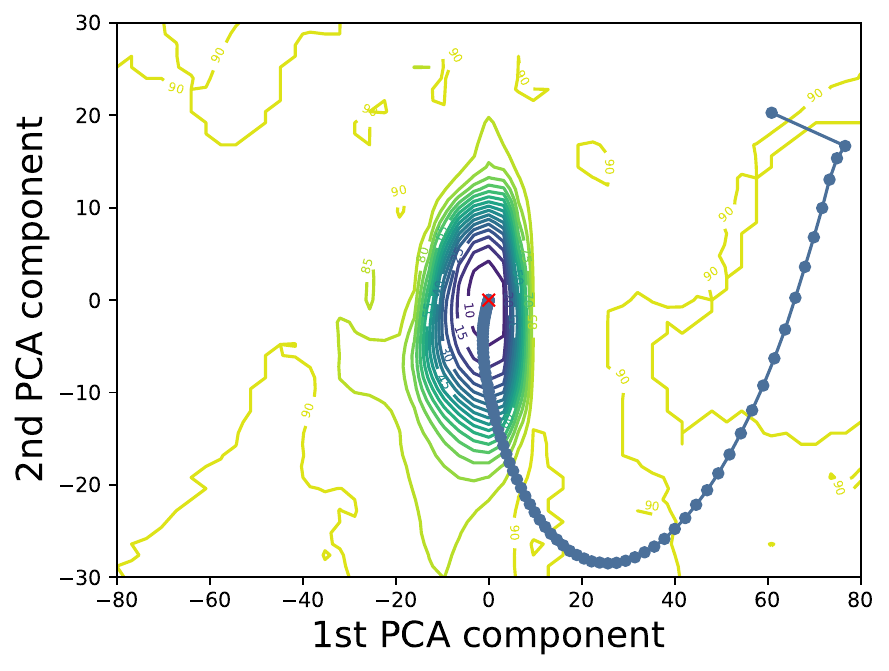}
		\caption{trajectory (supervised). left: R-20$\times$4, 94.07\%;
		right: R-20$\times$4$_{[1.0,0.5]}$,
		93.58\%, 93.03\%.}
		\label{fig:vis2}
	\end{subfigure}
	\begin{subfigure}{\linewidth}
		\centering
		\includegraphics[width=0.48\linewidth]{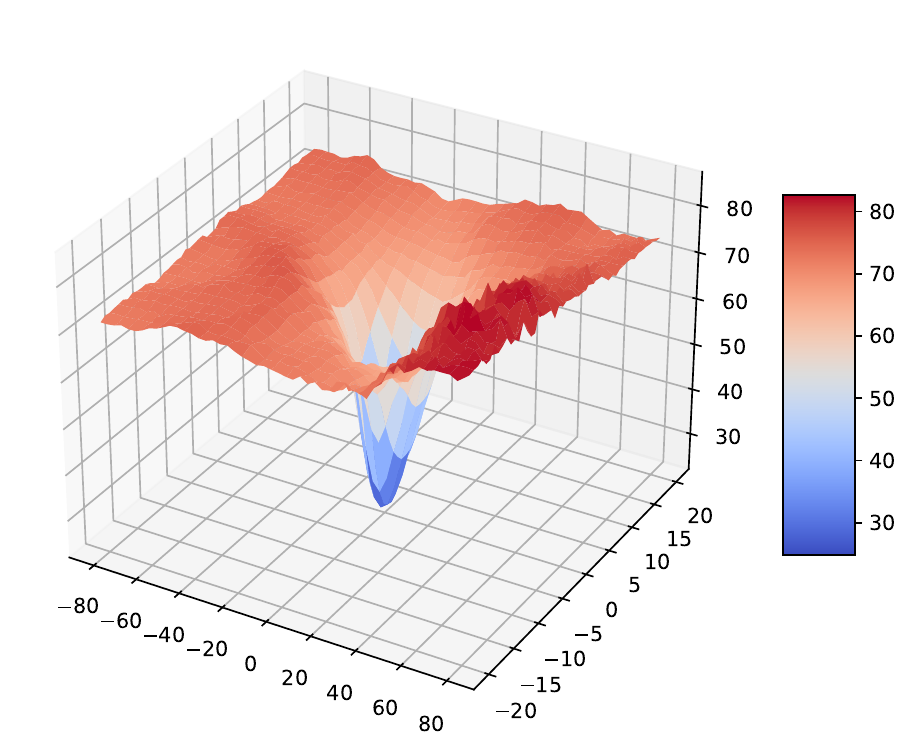}
		\includegraphics[width=0.48\linewidth]{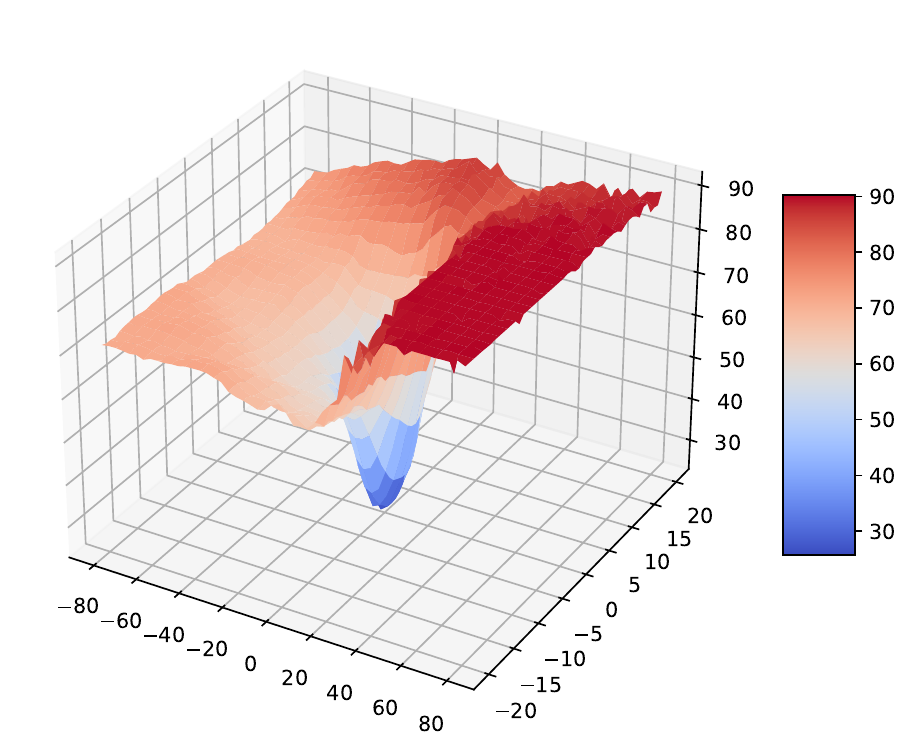}
		\caption{error surface (MoCo). left: R-20$\times$4, 76.75\%;
		right: R-20$\times$4$_{[1.0,0.5]}$,
		75.74\%, 74.24\%.}
		\label{fig:vis3}
	\end{subfigure}
	\begin{subfigure}{\linewidth}
		\centering
		\includegraphics[width=0.48\linewidth]{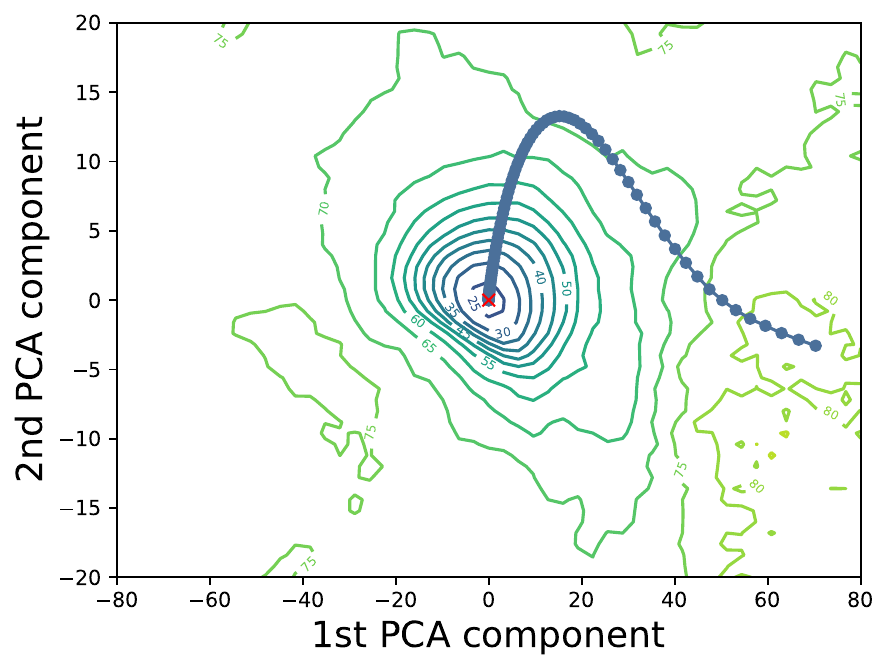}
		\includegraphics[width=0.48\linewidth]{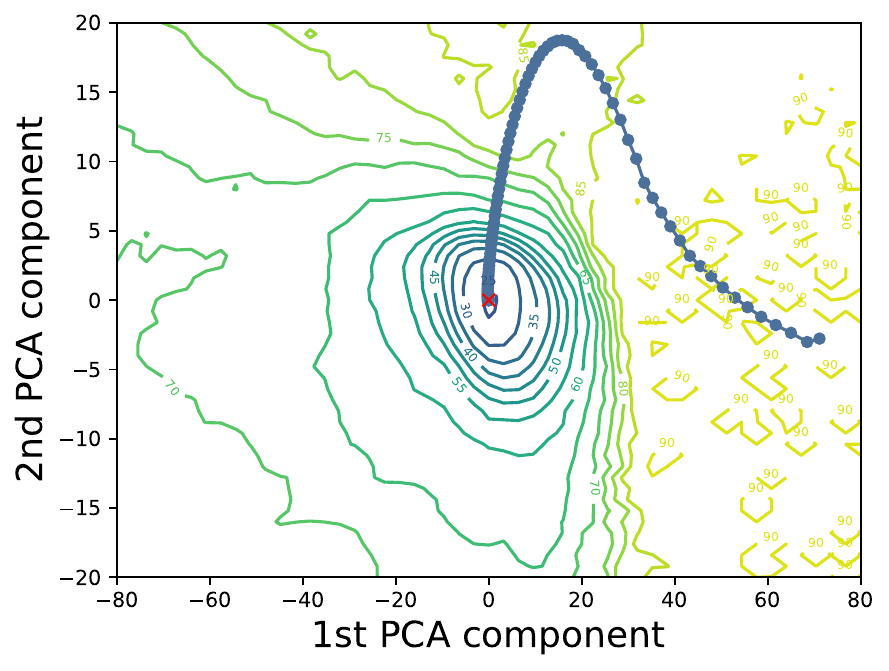}
		\caption{trajectory (MoCo). left: R-20$\times$4, 76.75\%;
		right: R-20$\times$4$_{[1.0,0.5]}$,
		75.74\%, 74.24\%.}
		\label{fig:vis4}
	\end{subfigure}
	\caption{Error surface and optimization trajectory. \textbf{Best viewed in color with 300\% zoom}.
	}
	\label{fig:vis}
\end{figure}

\noindent\textbf{Self-supervised and supervised slimmable networks~}
Training slimmable networks is generally more 
challenging in self-supervised learning than in 
supervised learning.
In Figure~\ref{fig3-grad} and Figure~\ref{fig-grad-diverge}, 
gradient imbalance and gradient direction divergence are more pronounced in self-supervised cases, leading to more severe 
performance degradation in Figure~\ref{fig1_scales}.
To gain a deeper understanding of the difficulty in training slimmable 
networks in self-supervised learning, we visualize the 
error surface and optimization 
trajectory~\citep{2018_vis} of slimmable networks 
during training.
Specifically, we train slimmable networks in both supervised and
self-supervised~(MoCo~\cite{2020_moco}) manners on CIFAR-
10~\citep{krizhevsky2009learning} for 100 epochs,
with a ResNet-20$\times$4 base network containing 4.3M parameters.
In self-supervised cases,
we use a $k$-NN predictor~\citep{wu2018unsupervised}
to obtain the accuracy.
After training, we visualize the error surface and optimization trajectory in Figure~\ref{fig:vis} following~\cite{2018_vis}.

The visualizations reveal that self-supervised learning is more challenging than supervised learning.
In the left error surface of Figure~\ref{fig:vis1} and Figure~\ref{fig:vis3}, it can be observed that the terrain surrounding the valley is relatively flat in supervised cases, while it is more complex in self-supervised cases.
Moreover, from the trajectory in the left of Figure~\ref{fig:vis2} and Figure~\ref{fig:vis4}, the contours in supervised cases are denser, indicating that the model in self-supervised cases requires more time to achieve the same accuracy improvement compared to the model in supervised cases.

The visualization indicates that weight-sharing networks have a 
greater impact in self-supervised cases.
Firstly, weight-sharing networks induce significant changes to the 
error surface in Figure~\ref{fig:vis3},
while the change is less apparent in supervised cases.
Secondly, in self-supervised cases, interference between weight-
sharing networks causes the model to deviate further from the 
global minima (\ie, the origin in the visualization) as 
illustrated in Figure~\ref{fig:vis4}.
In Figure~\ref{fig:vis2}, the maximal offsets from the global 
minima along the 2nd PCA component are 21.75 and 28.49 for ResNet-20$\times$4 and ResNet-20$\times$4$_{[1.0,0.5]}$, respectively. 
This represents a 31.0\% increase in offset.
Conversely, for self-supervised cases in Figure~\ref{fig:vis4}, 
the maximal offsets from the global minima along the 2nd PCA 
component are 13.26 and 18.75 for ResNet-20$\times$4 and ResNet-
20$\times$4$_{[1.0,0.5]}$, respectively. This increase in offset 
is 41.4\%. Therefore, it is evident that weight-sharing networks 
have a more pronounced impact in self-supervised cases.


By combining Figure~\ref{fig3-grad}~(gradient magnitude imbalance), Figure~\ref{fig-grad-diverge}~(gradient direction divergence), and Figure~\ref{fig:vis}~(optimization visualization), we posit a causal relationship between gradient magnitude imbalance, gradient direction divergence, and the challenge in training self-supervised slimmable networks.
The gradient magnitude imbalance and gradient direction divergence are evidences of interference between weight-sharing networks, making the training of slimmable networks more difficult than that of a normal network.
In the supervised case, consistence global ground truth alleviates the interference~(Figure~\ref{fig3:sub1}\&\ref{fig-gd:sub11}\&\ref{fig:vis1}\&\ref{fig:vis2}).
However, in the self-supervise case, the lack of consistent global ground truth makes the optimization of self-supervised slimmable networks harder than that of supervised slimmable networks~(Figure~\ref{fig3:sub2}\&\ref{fig-gd:sub12}\&\ref{fig:vis3}\&\ref{fig:vis4}).
During the pre-training stage of SlimCLR, we introduce online distillation, slow start and loss reweighting to mitigate the gradient magnitude imbalance and gradient direction divergence~(Figure~\ref{fig3:sub5}\&\ref{fig-gd:sub13}).
Additionally, to meet the conditions of inputs~(Appendix~\ref{app1}),  we propose the switchable linear probe layer for improved linear evaluation of a self-supervised slimmable network.

\section{Conclusion}
In this work,
we adapt slimmable networks for contrastive learning to obtain
pre-trained small models in a self-supervised manner.
By using slimmable networks, we can pre-train once and obtain several models of varying sizes, making them suitable for use across different devices.
Additionally, our approach does not require the additional training process of large teacher models, as seen in previous distillation-based methods.
However, weight-sharing networks in a slimmable network cause interference during self-supervised learning, resulting in gradient magnitude imbalance and gradient direction divergence.
We develop several techniques to relieve the interference among networks during pre-training and linear
evaluation.
Two specific algorithms are instantiated in this work,
\ie, SlimCLR-MoCov2 and SlimCLR-MoCov3.
We take extensive experiments on ImageNet and
achieve better performance than previous arts
with fewer network parameters and FLOPs.


%
\noindent\textbf{Conflict of interest.} The authors declare that they have no conflict of interest.

\noindent\textbf{Data availability.}
The datasets analysed during the current study are available in \url{https://www.image-net.org/},
\url{https://www.cs.toronto.edu/~kriz/cifar.html},
and \url{https://cocodataset.org/}.
No new datasets were generated.

\noindent\textbf{Acknowledgements.}
This work is supported by Major program of the National Natural Science Foundation of China (T2293720/T2293723). This work is also partially supported by Fundamental Research Funds for the Central Universities (Grant Number: 226-2022-00051).

\appendix
\section*{Appendix}
\section{Conditions of inputs} \label{app1}

We consider the conditions of inputs when only using
one slimmable linear  layer during evaluation, \ie, consider solving
multiple multi-class linear regression problems with shared weights.
The parameters of the linear layer are $\bm{\theta } \in \mathbb{R}^{d\times C}$,
$C$ is the number of classes, where $\bm{\theta } =\begin{bmatrix}
	\bm{\theta }_{11}\\
	\bm{\theta }_{21}
\end{bmatrix}$,
$\bm{\theta }_{11} \in \mathbb{R}^{d_{1} \times C}$,
$\bm{\theta }_{21} \in \mathbb{R}^{d_{2} \times C}$,
$d_{1} \ +\ d_{2} \ =\ d$.
The first input for the full model is $\bm{X} \in \mathbb{R}^{N\times d}$,
where $N$ is the number of samples, ${\bm{X}} =\begin{bmatrix}
	\bm{X}_{11} & \bm{X}_{12}
\end{bmatrix}$,
$\bm{X}_{11} \in \mathbb{R}^{N\times d_{1}}$,
$\bm{X}_{12} \ \in \mathbb{R}^{N\times d_{2}}$.
The second input $\bm{X}_{1} \in \mathbb{R}^{N\times d_{1}}$ is the input feature for 
the sub-model parameterized by $\bm{\theta }_{11}$.
Generally, we have $N\geq d >d_{1}$.
We assume that both $\bm{X}$ and $\bm{X}_{1}$ have independent columns, \ie, 
$\bm{X}^{T}\bm{X}$ and $\bm{X}_{1}^{T}\bm{X}_{1}$ are invertible.
The ground truth is $\bm{T} \in \mathbb{R}^{N\times C}$.
The prediction of the full model is $\bm{Y} = \bm{X} \bm{\theta }$,
to minimize the sum-of-least-squares loss between prediction and ground truth,
we get
\begin{align}
	\bm{\theta } =\argmin_{\bm{\theta}} \lVert\bm{X\theta } \ -\ \bm{T} \rVert_{2}^{2}.
\end{align}
By setting the derivative \textit{w.r.t.} $\bm{\theta }$ to 0, we get
\begin{align}
	\bm{\theta }  = \left(\bm{X}^{T}\bm{X}\right)^{-1}\bm{X}^{T}\bm{T}.
\end{align}
In the same way, we can get 
\begin{align}
	\bm{\theta }_{11}  = \left(\bm{X}_{1}^{T}\bm{X}_{1}\right)^{-1}\bm{X}_{1}^{T}\bm{T}.
\end{align}
For $\bm{X}^{T}\bm{X}$, we have
\begin{align}
	\bm{X}^{T}\bm{X}
 =\begin{bmatrix}
		\bm{X}_{11}^{T}\bm{X}_{11} & \bm{X}_{11}^{T}\bm{X}_{12}\\
		\bm{X}_{12}^{T} \bm{X}_{11} & \bm{X}_{12}^{T}\bm{X}_{12}
	\end{bmatrix}.
\end{align}
We denote the inverse of $\bm{X}^{T}\bm{X}$ as $\bm{B} =\begin{bmatrix}
	\bm{B}_{11} & \bm{B}_{12}\\
	\bm{B}_{21} & \bm{B}_{22}
\end{bmatrix}$,
where $\bm{B}_{12} =\bm{B}_{21}^{T}$
as $\bm{X}^{T}\bm{X}$ is a symmetric matrix.
For $\bm{X}^{T}\bm{XB} =\bm{I}$, we have
\begin{align}
	\bm{X}^{T}\bm{X} \begin{bmatrix}
		\bm{B}_{11} & \bm{B}_{12}\\
		\bm{B}_{21} & \bm{B}_{22}
	\end{bmatrix} =\begin{bmatrix}
		\bm{I}_{d_{1}} & \bm{0}_{d_{1} ,\ d_{2}}\\
		\bm{0}_{d_{2} ,\ d_{1}} & \bm{I}_{d_{2}}
	\end{bmatrix}.
\end{align}
Then we can get
\begin{align}
	\bm{X}_{11}^{T}\bm{X}_{11}\bm{B}_{11} +\bm{X}_{11}^{T}\bm{X}_{12}\bm{B}_{21} &=\bm{I}_{d_{1}},\label{frac_inv_1} \\
	\bm{X}_{11}^{T}\bm{X}_{11}\bm{B}_{12} +\bm{X}_{11}^{T}\bm{X}_{12}\bm{B}_{22} &=\bm{0}_{d_{1} ,\ d_{2}},\label{frac_inv_2} \\
	\bm{X}_{12}^{T}\bm{X}_{11}\bm{B}_{11} +\bm{X}_{12}^{T}\bm{X}_{12}\bm{B}_{21} &=\bm{0}_{d_{1} ,\ d_{2}}, \label{frac_inv_3} \\
	\bm{X}_{12}^{T}\bm{X}_{11}\bm{B}_{12} +\bm{X}_{12}^{T}\bm{X}_{12}\bm{B}_{22} &=\bm{I}_{d_{2}}. \label{frac_inv_4}
\end{align}
At the same time
\begin{align}
	\bm{\theta } &= \left(\bm{X}^{T}\bm{X}\right)^{-1}\bm{X}^{T}\bm{T}  =\bm{BX}^{T}\bm{T} \nonumber \\
	&=\begin{bmatrix}
		\bm{B}_{11} & \bm{B}_{12} \nonumber \\
		\bm{B}_{21} & \bm{B}_{22}
	\end{bmatrix}\begin{bmatrix}
		\bm{X}_{11} & \bm{X}_{12}
	\end{bmatrix}^{T}\bm{T} \\
	&=\begin{bmatrix}
		\bm{B}_{11}\bm{X}_{11}^{T} +\bm{B}_{12}\bm{X}_{12}^{T}\\
		\bm{B}_{21}\bm{X}_{11}^{T} +\bm{B}_{22}\bm{X}_{12}^{T}
	\end{bmatrix}\bm{T},
\end{align}
and
\begin{align} \label{thea_11}
	\bm{\theta }_{11} \ &=\ \left(\bm{B}_{11}\bm{X}_{11}^{T} +\bm{B}_{12}\bm{X}_{12}^{T}\right)\bm{T} \nonumber \\
 &=\left(\bm{X}_{1}^{T}\bm{X}_{1}\right)^{-1}\bm{X}_{1}^{T}\bm{T}.
\end{align}
From Eq.~\eqref{frac_inv_3}, we get 
\begin{align}
	\bm{B}_{21} &=-\left(\bm{X}_{12}^{T}\bm{X}_{12}\right)^{-1}\bm{X}_{12}^{T}\bm{X}_{11}\bm{B}_{11}, \label{eq_18}\\
	\bm{B}_{12} &=\bm{-B}_{11}\bm{X}_{11}^{T}\bm{X}_{12}\left(\bm{X}_{12}^{T}\bm{X}_{12}\right)^{-1}.
\end{align}
Substitute Eq.~\eqref{eq_18} into Eq.~\eqref{frac_inv_1}, we get
\begin{align}
	&\bm{B}_{11} \ = \nonumber \\
 &\left( \ \bm{X}_{11}^{T}\bm{X}_{11} -\bm{X}_{11}^{T}\bm{X}_{12}\left(\bm{X}_{12}^{T}\bm{X}_{12}\right)^{-1}\bm{X}_{12}^{T}\bm{X}_{11}\right)^{-1}.
\end{align}
At the same time
\begin{align} \label{theta_11_v2}
	\bm{\theta }_{11} &= \left(\bm{B}_{11}\bm{X}_{11}^{T} +\bm{B}_{12}\bm{X}_{12}^{T}\right)\bm{T} \notag\\
	&= \bm{B}_{11}\left(\bm{X}_{11}^{T} - \bm{X}_{11}^{T} \bm{X}_{12}\left( \bm{X}_{12}^{T} \bm{X}_{12}\right)^{-1} \bm{X}_{12}^{T}\right)\bm{T}.
\end{align}
Combining Eq.~\eqref{thea_11} and Eq.~\eqref{theta_11_v2}, we get the condition of the input
\begin{align} \label{eq_condition}
&\bm{B}_{11}\left(\bm{X}_{11}^{T} - \bm{X}_{11}^{T} \bm{X}_{12}\left( \bm{X}_{12}^{T} \bm{X}_{12}\right)^{-1} \bm{X}_{12}^{T}\right)\bm{T} =\notag\\ &\left(\bm{X}_{1}^{T}\bm{X}_{1}\right)^{-1}\bm{X}_{1}^{T}\bm{T}.
\end{align}

In order to verify whether the
input condition in Eq.~\eqref{eq_condition} 
is met in practice, we randomly sampled 2048 images from the training 
set of ImageNet and used a ResNet-50$_{[1.0, 0.75, 0.5, 0.25]}$ pre-
trained by SlimCLR-MoCov2~(800 epochs) to extract their features. The 
features extracted by ResNet-50$_{1.0}$ are denoted as $\bm{X} \in 
\mathbb{R}^{2048\times 1024}$, and the features extracted by
ResNet-50$_{0.5}$ are denoted as
$\bm{X}_1 \in \mathbb{R}^{2048\times 
512}$. We use $\bm{L}$ to represent the left side of 
Eq.~\eqref{eq_condition} and $\bm{R}$ for the right side. Then, we 
calculate the absolute difference between $\bm{L}$ and $\bm{R}$, 
denoted as $\lvert \bm{L} - \bm{R}\rvert$. The average value of 
entries in $\lvert \bm{L} - \bm{R}\rvert$ is $1.07$, which indicates 
a total difference of $1096665.50$. We also conducted similar 
experiments on the validation set of ImageNet and found that the 
average value of entries in $\lvert \bm{L} - \bm{R}\rvert$ is $0.88$, 
which indicates a total difference of $903094.19$.

These results demonstrate that the features of a slimmable network learned by contrastive 
self-supervised learning cannot meet the input 
conditions in Eq.~\eqref{eq_condition} when using a single 
slimmable linear probe layer. This provides an explanation for why 
using a switchable linear probe layer achieves much better 
performance in Table~\ref{tab2:sub1}.





\bibliographystyle{spbasic}      
\bibliography{ref}   

\begin{thebibliography}{65}
\providecommand{\natexlab}[1]{#1}
\providecommand{\url}[1]{{#1}}
\providecommand{\urlprefix}{URL }
\expandafter\ifx\csname urlstyle\endcsname\relax
  \providecommand{\doi}[1]{DOI~\discretionary{}{}{}#1}\else
  \providecommand{\doi}{DOI~\discretionary{}{}{}\begingroup
  \urlstyle{rm}\Url}\fi
\providecommand{\eprint}[2][]{\url{#2}}

\bibitem[{Bachman et~al.(2019)Bachman, Hjelm, and Buchwalter}]{2019_addim}
Bachman P, Hjelm RD, Buchwalter W (2019) Learning representations by maximizing
  mutual information across views. In: NeurIPS

\bibitem[{Ballard(1987)}]{1987_ae}
Ballard DH (1987) Modular learning in neural networks. In: AAAI

\bibitem[{Cai et~al.(2020)Cai, Gan, Wang, Zhang, and Han}]{2020_ofa}
Cai H, Gan C, Wang T, Zhang Z, Han S (2020) Once-for-all: Train one network and
  specialize it for efficient deployment. In: ICLR

\bibitem[{Caron et~al.(2018)Caron, Bojanowski, Joulin, and
  Douze}]{2018_deepcluster}
Caron M, Bojanowski P, Joulin A, Douze M (2018) Deep clustering for
  unsupervised learning of visual features. In: ECCV

\bibitem[{Caron et~al.(2020)Caron, Misra, Mairal, Goyal, Bojanowski, and
  Joulin}]{caron2020unsupervised}
Caron M, Misra I, Mairal J, Goyal P, Bojanowski P, Joulin A (2020) Unsupervised
  learning of visual features by contrasting cluster assignments. Advances in
  neural information processing systems 33:9912--9924

\bibitem[{Chavan et~al.(2022)Chavan, Shen, Liu, Liu, Cheng, and
  Xing}]{2022_vitslim}
Chavan A, Shen Z, Liu Z, Liu Z, Cheng KT, Xing EP (2022) Vision transformer
  slimming: Multi-dimension searching in continuous optimization space. In:
  CVPR

\bibitem[{Chen et~al.(2019)Chen, Wang, Pang, Cao, Xiong, Li, Sun, Feng, Liu,
  Xu, Zhang, Cheng, Zhu, Cheng, Zhao, Li, Lu, Zhu, Wu, Dai, Wang, Shi, Ouyang,
  Loy, and Lin}]{mmdetection}
Chen K, Wang J, Pang J, Cao Y, Xiong Y, Li X, Sun S, Feng W, Liu Z, Xu J, Zhang
  Z, Cheng D, Zhu C, Cheng T, Zhao Q, Li B, Lu X, Zhu R, Wu Y, Dai J, Wang J,
  Shi J, Ouyang W, Loy CC, Lin D (2019) {MMDetection}: Open mmlab detection
  toolbox and benchmark. arXiv preprint arXiv:190607155

\bibitem[{Chen et~al.(2020{\natexlab{a}})Chen, Kornblith, Norouzi, and
  Hinton}]{2020_simclr}
Chen T, Kornblith S, Norouzi M, Hinton GE (2020{\natexlab{a}}) A simple
  framework for contrastive learning of visual representations. In: {ICML}

\bibitem[{Chen et~al.(2020{\natexlab{b}})Chen, Kornblith, Swersky, Norouzi, and
  Hinton}]{2020_simclrv2}
Chen T, Kornblith S, Swersky K, Norouzi M, Hinton GE (2020{\natexlab{b}}) Big
  self-supervised models are strong semi-supervised learners. In: NeurIPS

\bibitem[{Chen and He(2021)}]{2021_simsiam}
Chen X, He K (2021) Exploring simple siamese representation learning. In: CVPR

\bibitem[{Chen et~al.(2020{\natexlab{c}})Chen, Fan, Girshick, and He}]{mocov2}
Chen X, Fan H, Girshick RB, He K (2020{\natexlab{c}}) Improved baselines with
  momentum contrastive learning. CoRR abs/2003.04297, \eprint{2003.04297}

\bibitem[{Chen et~al.(2021)Chen, Xie, and He}]{2021_mocov3}
Chen X, Xie S, He K (2021) An empirical study of training self-supervised
  vision transformers. In: ICCV

\bibitem[{Devlin et~al.(2019)Devlin, Chang, Lee, and Toutanova}]{2019_bert}
Devlin J, Chang M, Lee K, Toutanova K (2019) {BERT:} pre-training of deep
  bidirectional transformers for language understanding. In: NAACL-HLT

\bibitem[{Dosovitskiy et~al.(2016)Dosovitskiy, Fischer, Springenberg,
  Riedmiller, and Brox}]{DosovitskiyFSRB16}
Dosovitskiy A, Fischer P, Springenberg JT, Riedmiller MA, Brox T (2016)
  Discriminative unsupervised feature learning with exemplar convolutional
  neural networks. TPAMI

\bibitem[{Fang et~al.(2021)Fang, Wang, Wang, Zhang, Yang, and Liu}]{seed}
Fang Z, Wang J, Wang L, Zhang L, Yang Y, Liu Z (2021) {SEED:} self-supervised
  distillation for visual representation. In: {ICLR}

\bibitem[{Gao et~al.(2022)Gao, Zhuang, Lin, Cheng, Sun, Li, and Shen}]{disco}
Gao Y, Zhuang J, Lin S, Cheng H, Sun X, Li K, Shen C (2022) Disco: Remedy
  self-supervised learning on lightweight models with distilled contrastive
  learning. In: ECCV

\bibitem[{Goyal et~al.(2017)Goyal, Doll{\'{a}}r, Girshick, Noordhuis,
  Wesolowski, Kyrola, Tulloch, Jia, and He}]{2017_PriyaGoyal}
Goyal P, Doll{\'{a}}r P, Girshick RB, Noordhuis P, Wesolowski L, Kyrola A,
  Tulloch A, Jia Y, He K (2017) Accurate, large minibatch {SGD:} training
  imagenet in 1 hour. CoRR abs/1706.02677, \eprint{1706.02677}

\bibitem[{Gu et~al.(2021)Gu, Liu, and Tian}]{gu2021simple}
Gu J, Liu W, Tian Y (2021) Simple distillation baselines for improving small
  self-supervised models. arXiv preprint arXiv:210611304

\bibitem[{Guo(2022)}]{guo2022reducing}
Guo J (2022) Reducing the teacher-student gap via adaptive temperatures

\bibitem[{He et~al.(2016)He, Zhang, Ren, and Sun}]{2016_ResNet}
He K, Zhang X, Ren S, Sun J (2016) Deep residual learning for image
  recognition. In: CVPR

\bibitem[{He et~al.(2017)He, Gkioxari, Doll{\'{a}}r, and
  Girshick}]{2017_maskrcnn}
He K, Gkioxari G, Doll{\'{a}}r P, Girshick RB (2017) Mask {R-CNN}. In: ICCV, pp
  2980--2988

\bibitem[{He et~al.(2020)He, Fan, Wu, Xie, and Girshick}]{2020_moco}
He K, Fan H, Wu Y, Xie S, Girshick RB (2020) Momentum contrast for unsupervised
  visual representation learning. In: CVPR

\bibitem[{He et~al.(2022)He, Chen, Xie, Li, Doll{\'a}r, and
  Girshick}]{2022_mae}
He K, Chen X, Xie S, Li Y, Doll{\'a}r P, Girshick R (2022) Masked autoencoders
  are scalable vision learners. In: CVPR

\bibitem[{He et~al.(2019)He, Zhang, Zhang, Zhang, Xie, and
  Li}]{2018_bags_of_tricks}
He T, Zhang Z, Zhang H, Zhang Z, Xie J, Li M (2019) Bag of tricks for image
  classification with convolutional neural networks. In: CVPR

\bibitem[{Hinton et~al.(2015)Hinton, Vinyals, and Dean}]{2015_kd}
Hinton GE, Vinyals O, Dean J (2015) Distilling the knowledge in a neural
  network. CoRR abs/1503.02531

\bibitem[{Hjelm et~al.(2019)Hjelm, Fedorov, Lavoie{-}Marchildon, Grewal,
  Bachman, Trischler, and Bengio}]{2019_infomax}
Hjelm RD, Fedorov A, Lavoie{-}Marchildon S, Grewal K, Bachman P, Trischler A,
  Bengio Y (2019) Learning deep representations by mutual information
  estimation and maximization. In: ICLR

\bibitem[{Ioffe and Szegedy(2015)}]{2015_bn}
Ioffe S, Szegedy C (2015) Batch normalization: Accelerating deep network
  training by reducing internal covariate shift. In: ICML

\bibitem[{Kingma and Welling(2014)}]{2014_VAE}
Kingma DP, Welling M (2014) Auto-encoding variational bayes. In: ICLR

\bibitem[{Koohpayegani et~al.(2020)Koohpayegani, Tejankar, and
  Pirsiavash}]{compress}
Koohpayegani SA, Tejankar A, Pirsiavash H (2020) Compress: Self-supervised
  learning by compressing representations. In: NeurIPS

\bibitem[{Krizhevsky and Hinton(2009)}]{krizhevsky2009learning}
Krizhevsky A, Hinton G (2009) Learning multiple layers of features from tiny
  images. Technical report, University of Toronto

\bibitem[{Li et~al.(2021)Li, Wang, Wang, Liang, Li, and Chang}]{Li_2021_CVPR}
Li C, Wang G, Wang B, Liang X, Li Z, Chang X (2021) Dynamic slimmable network.
  In: CVPR

\bibitem[{Li et~al.(2022)Li, Wang, Wang, Liang, Li, and Chang}]{Li_2022_pami}
Li C, Wang G, Wang B, Liang X, Li Z, Chang X (2022) Ds-net++: Dynamic weight
  slicing for efficient inference in cnns and transformers. T-PAMI

\bibitem[{Li et~al.(2018)Li, Xu, Taylor, Studer, and Goldstein}]{2018_vis}
Li H, Xu Z, Taylor G, Studer C, Goldstein T (2018) Visualizing the loss
  landscape of neural nets. In: NeurIPS

\bibitem[{Lin et~al.(2014)Lin, Maire, Belongie, Hays, Perona, Ramanan,
  Doll{\'{a}}r, and Zitnick}]{2014_coco}
Lin T, Maire M, Belongie SJ, Hays J, Perona P, Ramanan D, Doll{\'{a}}r P,
  Zitnick CL (2014) Microsoft {COCO:} common objects in context. In: ECCV

\bibitem[{Lin et~al.(2017)Lin, Doll{\'{a}}r, Girshick, He, Hariharan, and
  Belongie}]{2017_fpn}
Lin T, Doll{\'{a}}r P, Girshick RB, He K, Hariharan B, Belongie SJ (2017)
  Feature pyramid networks for object detection. In: CVPR, pp 936--944

\bibitem[{Liu et~al.(2020)Liu, Zhang, Hou, Wang, Mian, Zhang, and
  Tang}]{2021_survey}
Liu X, Zhang F, Hou Z, Wang Z, Mian L, Zhang J, Tang J (2020) Self-supervised
  learning: Generative or contrastive. CoRR abs/2006.08218

\bibitem[{Liu et~al.(2017)Liu, Li, Shen, Huang, Yan, and
  Zhang}]{2017_networkslim}
Liu Z, Li J, Shen Z, Huang G, Yan S, Zhang C (2017) Learning efficient
  convolutional networks through network slimming. In: ICCV

\bibitem[{Micikevicius et~al.(2018)Micikevicius, Narang, Alben, Diamos, Elsen,
  Garc{\'{\i}}a, Ginsburg, Houston, Kuchaiev, Venkatesh, and Wu}]{2018_AMP}
Micikevicius P, Narang S, Alben J, Diamos GF, Elsen E, Garc{\'{\i}}a D,
  Ginsburg B, Houston M, Kuchaiev O, Venkatesh G, Wu H (2018) Mixed precision
  training. In: ICLR

\bibitem[{Mirzadeh et~al.(2020)Mirzadeh, Farajtabar, Li, Levine, Matsukawa, and
  Ghasemzadeh}]{2020_improved_kd}
Mirzadeh S, Farajtabar M, Li A, Levine N, Matsukawa A, Ghasemzadeh H (2020)
  Improved knowledge distillation via teacher assistant. In: AAAI

\bibitem[{NVIDIA(2021)}]{dali}
NVIDIA (2021) Nvidia dali: The nvidia data loading library.
  \urlprefix\url{https://docs.nvidia.com/deeplearning/dali/user-guide/docs/index.html}

\bibitem[{van~den Oord et~al.(2016{\natexlab{a}})van~den Oord, Kalchbrenner,
  Espeholt, Kavukcuoglu, Vinyals, and Graves}]{2016_pixelcnn}
van~den Oord A, Kalchbrenner N, Espeholt L, Kavukcuoglu K, Vinyals O, Graves A
  (2016{\natexlab{a}}) Conditional image generation with pixelcnn decoders. In:
  NeurIPS

\bibitem[{van~den Oord et~al.(2016{\natexlab{b}})van~den Oord, Kalchbrenner,
  and Kavukcuoglu}]{2016_pixelrnn}
van~den Oord A, Kalchbrenner N, Kavukcuoglu K (2016{\natexlab{b}}) Pixel
  recurrent neural networks. In: ICML

\bibitem[{van~den Oord et~al.(2018)van~den Oord, Li, and Vinyals}]{cpcv1}
van~den Oord A, Li Y, Vinyals O (2018) Representation learning with contrastive
  predictive coding. CoRR abs/1807.03748

\bibitem[{Park et~al.(2019)Park, Kim, Lu, and Cho}]{2019_relationalkd}
Park W, Kim D, Lu Y, Cho M (2019) Relational knowledge distillation. In: CVPR,
  \doi{10.1109/CVPR.2019.00409}

\bibitem[{Pathak et~al.(2016)Pathak, Kr{\"{a}}henb{\"{u}}hl, Donahue, Darrell,
  and Efros}]{2016_inpaint}
Pathak D, Kr{\"{a}}henb{\"{u}}hl P, Donahue J, Darrell T, Efros AA (2016)
  Context encoders: Feature learning by inpainting. In: CVPR

\bibitem[{Peng et~al.(2019)Peng, Jin, Li, Zhou, Wu, Liu, Zhang, and
  Liu}]{2019_CCKD}
Peng B, Jin X, Li D, Zhou S, Wu Y, Liu J, Zhang Z, Liu Y (2019) Correlation
  congruence for knowledge distillation. In: ICCV,
  \doi{10.1109/ICCV.2019.00511}

\bibitem[{Radford et~al.(2016)Radford, Metz, and Chintala}]{2016_dcgans}
Radford A, Metz L, Chintala S (2016) Unsupervised representation learning with
  deep convolutional generative adversarial networks. In: ICLR

\bibitem[{Russakovsky et~al.(2015)Russakovsky, Deng, Su, Krause, Satheesh, Ma,
  Huang, Karpathy, Khosla, Bernstein, Berg, and Li}]{ImageNet}
Russakovsky O, Deng J, Su H, Krause J, Satheesh S, Ma S, Huang Z, Karpathy A,
  Khosla A, Bernstein MS, Berg AC, Li F (2015) Imagenet large scale visual
  recognition challenge. IJCV

\bibitem[{Shi et~al.(2022)Shi, Zhang, Tang, Zhu, Li, Guo, and
  Zhuang}]{shi2022efficacy}
Shi H, Zhang Y, Tang S, Zhu W, Li Y, Guo Y, Zhuang Y (2022) On the efficacy of
  small self-supervised contrastive models without distillation signals. In:
  AAAI, vol~36, pp 2225--2234

\bibitem[{Tan and Le(2019)}]{tan2019efficientnet}
Tan M, Le Q (2019) Efficientnet: Rethinking model scaling for convolutional
  neural networks. In: ICML, PMLR, pp 6105--6114

\bibitem[{Tian et~al.(2020)Tian, Krishnan, and Isola}]{2020_CRD}
Tian Y, Krishnan D, Isola P (2020) Contrastive representation distillation. In:
  ICLR

\bibitem[{Tomasev et~al.(2022)Tomasev, Bica, McWilliams, Buesing, Pascanu,
  Blundell, and Mitrovic}]{relicv2}
Tomasev N, Bica I, McWilliams B, Buesing L, Pascanu R, Blundell C, Mitrovic J
  (2022) Pushing the limits of self-supervised resnets: Can we outperform
  supervised learning without labels on imagenet? CoRR abs/2201.05119

\bibitem[{Wang et~al.(2021)Wang, Chen, Zhao, Chen, Hu, Liu, Cai, He, and
  Liu}]{pmlr-v139-wang21e}
Wang W, Chen M, Zhao S, Chen L, Hu J, Liu H, Cai D, He X, Liu W (2021)
  Accelerate cnns from three dimensions: A comprehensive pruning framework. In:
  ICML

\bibitem[{Wu et~al.(2018)Wu, Xiong, Yu, and Lin}]{wu2018unsupervised}
Wu Z, Xiong Y, Yu SX, Lin D (2018) Unsupervised feature learning via
  non-parametric instance discrimination. In: CVPR

\bibitem[{Xu et~al.(2022)Xu, Fang, Zhang, Xie, Wang, Dai, Xiong, and
  Tian}]{2022_bingo}
Xu H, Fang J, Zhang X, Xie L, Wang X, Dai W, Xiong H, Tian Q (2022) Bag of
  instances aggregation boosts self-supervised distillation. In: ICLR

\bibitem[{Yang et~al.(2020)Yang, Zhu, Chen, Yan, Zhang, and
  Willis}]{2020_mutualnet}
Yang T, Zhu S, Chen C, Yan S, Zhang M, Willis AR (2020) Mutualnet: Adaptive
  convnet via mutual learning from network width and resolution. In: ECCV

\bibitem[{You et~al.(2017)You, Gitman, and Ginsburg}]{you2017large}
You Y, Gitman I, Ginsburg B (2017) Large batch training of convolutional
  networks. arXiv preprint arXiv:170803888

\bibitem[{Yu and Huang(2019{\natexlab{a}})}]{yu2019autoslim}
Yu J, Huang T (2019{\natexlab{a}}) Autoslim: Towards one-shot architecture
  search for channel numbers. arXiv preprint arXiv:190311728

\bibitem[{Yu and Huang(2019{\natexlab{b}})}]{2019_unslim}
Yu J, Huang TS (2019{\natexlab{b}}) Universally slimmable networks and improved
  training techniques. In: ICCV

\bibitem[{Yu et~al.(2019)Yu, Yang, Xu, Yang, and Huang}]{2019_slim}
Yu J, Yang L, Xu N, Yang J, Huang TS (2019) Slimmable neural networks. In: ICLR

\bibitem[{Yu et~al.(2020)Yu, Jin, Liu, Bender, Kindermans, Tan, Huang, Song,
  Pang, and Le}]{2020_bignas}
Yu J, Jin P, Liu H, Bender G, Kindermans P, Tan M, Huang TS, Song X, Pang R, Le
  Q (2020) Bignas: Scaling up neural architecture search with big single-stage
  models. In: ECCV, Springer, pp 702--717

\bibitem[{Zhang et~al.(2016)Zhang, Isola, and Efros}]{2016_color}
Zhang R, Isola P, Efros AA (2016) Colorful image colorization. In: ECCV

\bibitem[{Zhang et~al.(2018)Zhang, Xiang, Hospedales, and Lu}]{2018_dml}
Zhang Y, Xiang T, Hospedales TM, Lu H (2018) Deep mutual learning. In: CVPR,
  \doi{10.1109/CVPR.2018.00454}

\bibitem[{Zhao et~al.(2022{\natexlab{a}})Zhao, Cui, Song, Qiu, and
  Liang}]{zhao2022decoupled}
Zhao B, Cui Q, Song R, Qiu Y, Liang J (2022{\natexlab{a}}) Decoupled knowledge
  distillation. In: CVPR

\bibitem[{Zhao et~al.(2022{\natexlab{b}})Zhao, Zhou, Wang, Cai, Lam, and
  Xu}]{2020_dc}
Zhao S, Zhou L, Wang W, Cai D, Lam TL, Xu Y (2022{\natexlab{b}}) Toward better
  accuracy-efficiency trade-offs: Divide and co-training. TIP 31:5869--5880

\end{thebibliography}

%
%

\end{document}